\documentclass[lettersize,journal]{IEEEtran}
\usepackage{amsmath,amsfonts}
\usepackage{algorithmic}
\usepackage{algorithm}
\usepackage{array}
\usepackage[caption=false,font=normalsize,labelfont=sf,textfont=sf]{subfig}
\usepackage{textcomp}
\usepackage{stfloats}
\usepackage{url}
\usepackage{verbatim}
\usepackage{graphicx}
\usepackage{cite}
\usepackage{amssymb}
\usepackage{hyperref}
\usepackage{wrapfig}
\usepackage{booktabs}
\usepackage{soul}
\usepackage{color}
\usepackage{amsthm}
\usepackage{mathtools}
\usepackage{multirow}
\usepackage{booktabs}
\usepackage{bbding}
\usepackage{cuted}    
\usepackage{capt-of}  

\begin{document}

\title{\includegraphics[height=1.8em]{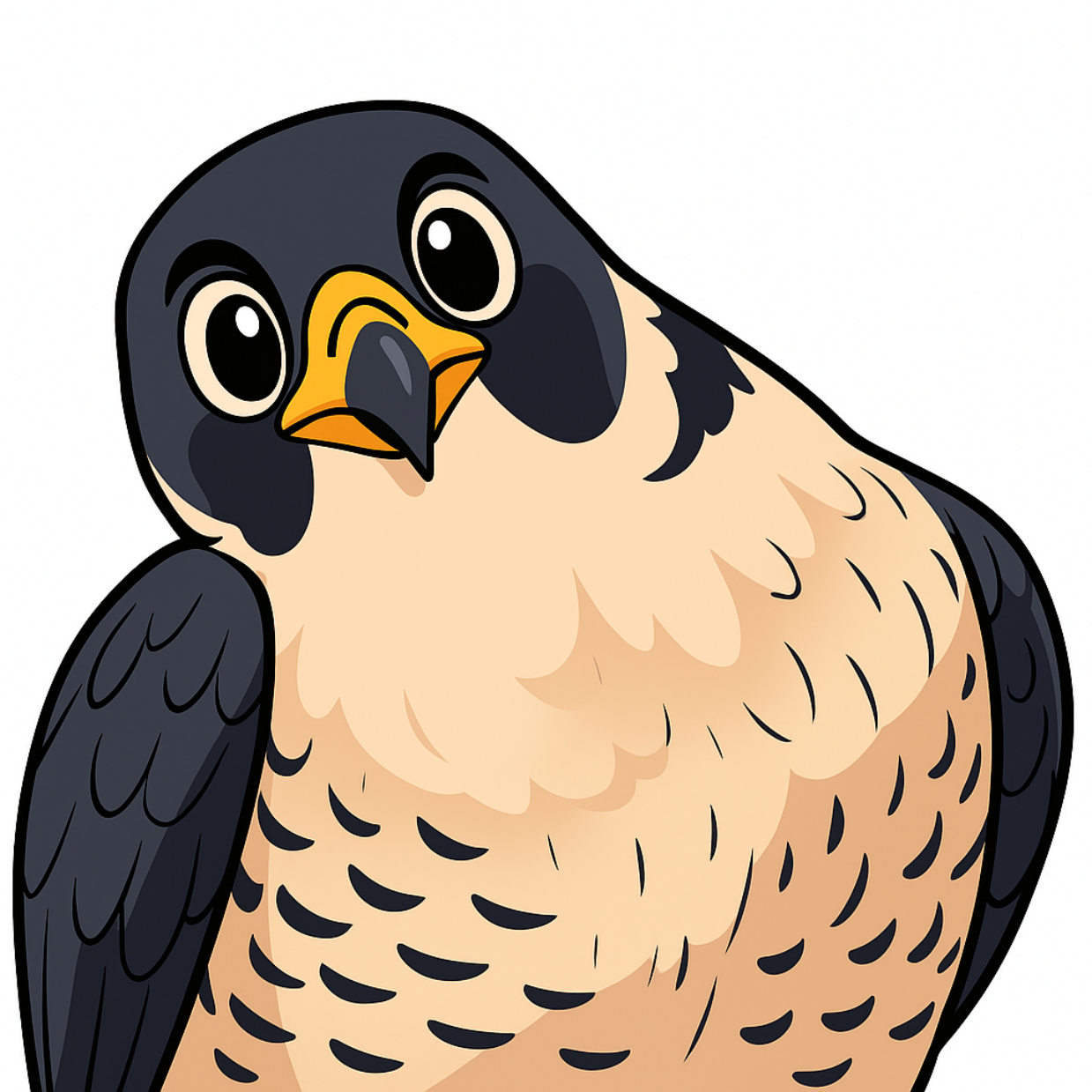}FALCON: Actively Decoupled Visuomotor Policies for Loco-Manipulation with Foundation-Model-Based Coordination
\\ {\large Project Page: \url{https://marmotlab.github.io/falcon/}}}

\author{Chengyang He$^{1,*}$, Ge Sun$^{1,*,\dagger}$\thanks{$\dagger$ Corresponding author: sunge@u.nus.edu}, Yue Bai$^{1}$, Junkai Lu$^{1}$, Jiadong Zhao$^{1}$, Guillaume Sartoretti$^{1}$
\thanks{$^{*}$ Equal Contribution.}
\thanks{$^1$ Department of Mechanical Engineering, College of Design and Engineering, National University of Singapore.}}

\markboth{Journal of \LaTeX\ Class Files,~Vol.~*, No.~*, November~2025}%
{Shell \MakeLowercase{\textit{et al.}}: A Sample Article Using IEEEtran.cls for IEEE Journals}


\maketitle
\begin{strip}
    \centering
    \includegraphics[width=\textwidth]{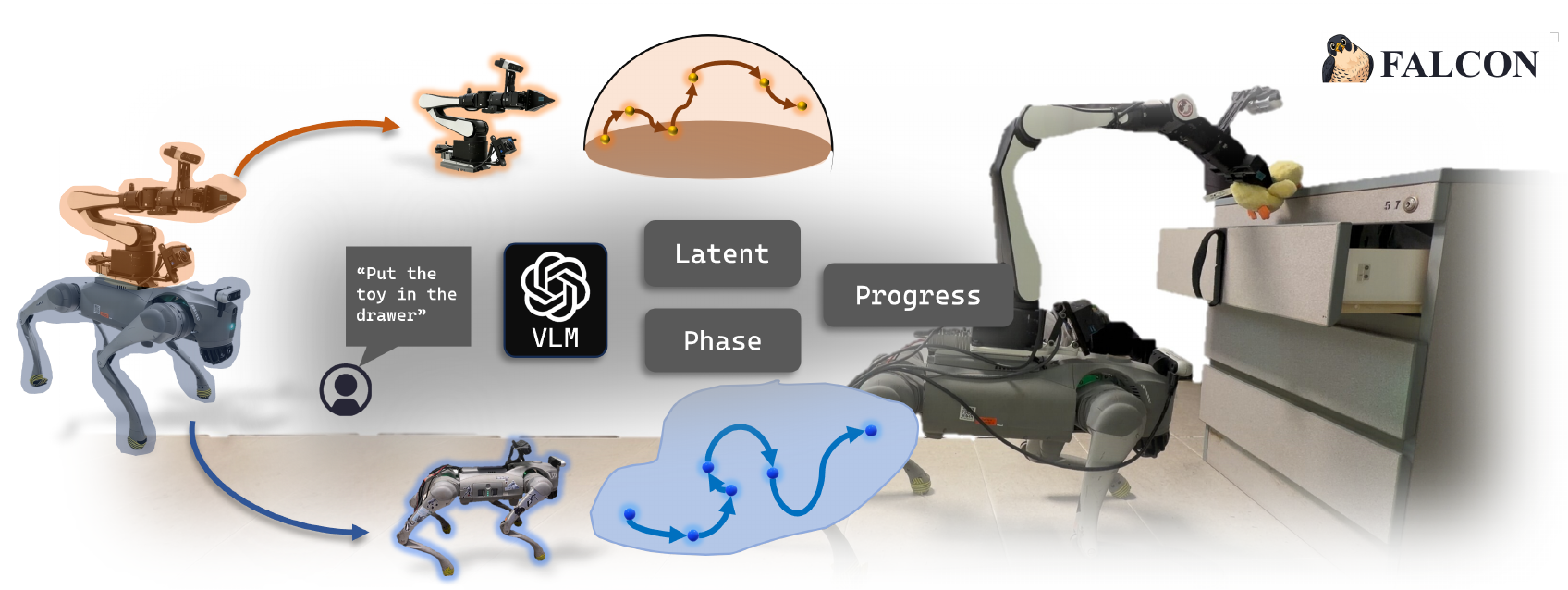}
    \captionof{figure}{\textbf{FALCON} actively decouples locomotion and manipulation through two modular diffusion policies, coordinated by a vision-language foundation model. The VLM encodes global scene context, proprioceptive states, and goal instructions into a shared latent embedding that conditions both subsystems. }
    \label{fig:banner_long}
\end{strip}
\vspace{-1em}

\begin{abstract}

We present FoundAtion-model-guided decoupled LoCO-maNipulation visuomotor policies (FALCON), a framework for loco-manipulation that combines modular diffusion policies with a vision–language foundation model as the coordinator. Our approach explicitly decouples locomotion and manipulation into two specialized visuomotor policies, allowing each subsystem to rely on its own observations.  This mitigates the performance degradation that arise when a single policy is forced to fuse heterogeneous, potentially mismatched observations from locomotion and manipulation. Our key innovation lies in restoring coordination between these two independent policies through a vision–language foundation model, which encodes global observations and language instructions into a shared latent embedding conditioning both diffusion policies. On top of this backbone, we introduce a phase-progress head that uses textual descriptions of task stages to infer discrete phase and continuous progress estimates without manual phase labels. To further structure the latent space, we incorporate a coordination-aware contrastive loss that explicitly encodes cross-subsystem compatibility between arm and base actions. We evaluate FALCON on two challenging loco-manipulation tasks requiring navigation, precise end-effector placement, and tight base-arm coordination.  Results show that it surpasses centralized and decentralized baselines while exhibiting improved robustness and generalization to out-of-distribution scenarios.

\end{abstract}

\begin{IEEEkeywords}
Foundation Model for Robotics, Visuomotor Policy, Decoupled Decision Making, Loco-Manipulation, Contrastive Learning
\end{IEEEkeywords}

\section{Introduction}\label{sec:intro}
\IEEEPARstart{R}{ecent} progress in robot learning and foundation models has rekindled the longstanding vision of general-purpose robots that can move through unstructured environments and manipulate diverse objects with minimal task-specific engineering. 
Large Behavior Models (LBMs) extend the diffusion policy paradigm to multi-task dexterous manipulation~\cite{barreiros2025careful}, training a single policy across broad datasets of real and simulated trajectories.
At the same time, several industrial efforts, such as Boston Dynamics and Toyota Research Institute's Atlas LBM~\cite{BostonDynamics_Atlas2025}, Figure’s Helix vision-language-action (VLA) model~\cite{FigureAI_Helix2025}, NVIDIA’s GR00T N1.5~\cite{bjorck2025gr00t}, Physical Intelligence’s $\pi_{0/0.5}$~\cite{black2024pi0,intelligence2025pi05}, Flexion's Reflect~\cite{Flexion_Reflect2025}, and Sunday Robotics’ Memo platform~\cite{SundayRobotics_Technology2025}, demonstrate impressive whole-body behaviors that combine locomotion, manipulation, and language grounding in increasingly realistic environments.
These developments suggest a future where robot generalist models consume raw sensor streams and language instructions and directly output actions to interact with the physical world. 

However, loco-manipulation, jointly controlling a mobile base and one or more arms, remains especially challenging on legged platforms~\cite{zhu2025versatile,sleiman2023versatile,gu2025humanoid}, where the same body must simultaneously maintain stability and accomplish precise manipulation under different sensor streams and poses. 
In this work, we focus on a specific yet representative setting in which an arm-mounted quadruped robot performs long-horizon loco-manipulation tasks using only RGB observations, proprioceptive states, and sparse language instructions.
Within this setting, we revisit several design choices that are often implicit in vision-language model (VLM)-based loco-manipulation architectures and argue for three key departures: (i) deliberately \emph{decoupling} locomotion and manipulation control instead of using a single whole-body policy, (ii) using a vision-language foundation model not as an end-to-end controller, but as a shared semantic coordinator between specialized diffusion policies, and (iii) providing both task-level language prompts and carefully designed, phase-specific textual descriptions, allowing the VLM to produce temporally grounded phase and progress.

\textbf{\textit{Why legged locomotion rather than wheeled motion?}}
Most deployed mobile manipulators today are wheeled platforms, such as Sunday Robotics’ Memo, Physical Intelligence's $\pi_0$, and Stanford's Mobile ALOHA, which benefit from mature navigation stacks and relatively simple dynamics.
However, wheeled robots are fundamentally constrained to relatively smooth, structured environments, like warehouses, conference venues, and shopping malls.
Legged mobile manipulators, by contrast, can traverse stairs, clutter, and uneven terrain while repositioning the base to improve manipulability~\cite{rehman2016towards}. 
This makes them attractive candidates for the kind of household and industrial environments envisioned by generalist robotics efforts.
But this choice also significantly complicates loco-manipulation: the same body must satisfy both dynamic stability constraints and manipulation objectives, and errors in base control can quickly degrade manipulation performance.
Our work builds on this line but focuses on a more constrained yet practically relevant regime: an manipulator-mounted quadruped operating with RGB images only, and with fixed limb roles (legs for locomotion, arm for manipulation). 
This setting emphasizes the challenges of (i) long-horizon navigation with limited perception, (ii) precise end-effector placement from 2D images, and (iii) robust coordination between base motion and arm manipulation under partial observability.

\textbf{\textit{Why decouple locomotion and manipulation?}}
Most whole-body control approaches, both classical and learning-based, seek a single policy that directly maps joint- or task-space observations to all robot actuators. 
Examples range from model-based whole-body MPC for legged mobile manipulators to deep whole-body policies that jointly optimize base and arm motion.
While such monolithic policies can produce elegant, tightly coordinated motions, they also suffer from several drawbacks, including difficulty in training, sensitivity to dataset coverage, and a tendency to fail when one subsystem encounters out-of-distribution conditions (e.g., unexpected terrain) that corrupt the shared latent representation.
Recent work on Latent Theory of Mind (LatentToM)~\cite{he2025latent} shows that, in pure manipulation settings, decentralizing control and explicitly structuring latent spaces into ego and consensus components can improve robustness and coordination among multiple arms.
In loco-manipulation, the asymmetry between the base and the manipulator is even more pronounced, as a quadruped must handle dynamic stability and long-horizon navigation, often under open-loop velocity commands, while the manipulator executes precise position commands through inverse kinematics.
Treating them as a single homogeneous control problem forces the policy to fuse heterogeneous observations and control objectives, amplifying optimization and generalization difficulties.

In contrast, we propose to \emph{actively decouple} the system into two diffusion-based visuomotor policies: a high-level velocity policy for locomotion on top of a pre-trained low-level RL joint controller, and a high-level end-effector position policy for the arm. 
Each policy operates on its own observation and action space.
Therefore,  this decoupled regime allows one human take over part of it easily when facing any confusing cases.
Additionally, this modularity reduces the risk of observation mismatch, allowing each subsystem to adopt its most natural control regime. 
The resulting challenge is no longer ``how to learn one gigantic whole-body policy,'' but rather ``how to coordinate two strong but independent specialists.''

\textbf{\textit{How can foundation model help with coordination?}}
Foundation models for robotics, such as $\pi_0$, GR00T N1, Helix, and LBMs, typically adopt an end-to-end vision-language-action formulation in which a large vision-language backbone encodes images and text while a flow-matching or diffusion-based action decoder outputs whole-body motor commands.
While these models achieve remarkable generality across tasks and embodiments, they also demand massive, carefully curated datasets and significant compute, and their internal coordination mechanisms are opaque and hard to control.
Our design takes a different stance by treating the foundation model not as the policy itself but as a \emph{semantic coordinator} between otherwise decoupled diffusion policies.
Concretely, FALCON leverages CLIP~\cite{radford2021learning}, a widely used vision-language model pre-trained on internet-scale image–text pairs—to encode global RGB observations, proprioceptive robot states, and natural language task specifications into a shared latent embedding. 
This embedding is then injected as part of the conditioning input to both the locomotion and manipulation diffusion policies.
This design offers several advantages over fully end-to-end VLA controllers. 
First, CLIP is frozen and used purely as a feature extractor, avoiding catastrophic interference between foundation model pretraining and task-specific visuomotor learning. 
Second, this shared embedding provides a common ``task and scene summary'' that aligns the two policies without forcing them to share raw observations or low-level features with each other. 
Third, because coordination is mediated through a compact embedding, our approach is compatible with both centralized and decentralized execution, as the locomotion and manipulation policies can in principle run on separate compute units while still being conditioned on the same CLIP generated embedding.
Compared to LatentToM, which learns a consensus representation by sheaf-theoretic consistency losses from scratch, our method leverages an existing foundation model as the consensus channel, which can simplify training.

We present FoundAtion-model-guided decoupled LoCO-maNipulation visuomotor policies (FALCON), a framework that combines modular diffusion policies with a vision-language foundation model to address the above challenges. 
FALCON is built around three design principles:
\begin{itemize}
    \item \textbf{Decoupled control for loco-manipulation.} Instead of a monolithic whole-body policy, FALCON decouples locomotion and manipulation into independent visuomotor specialists.
    Each subsystem is trained in its own observation and control space, avoiding the observation mismatch failure modes common in whole-body controllers, while coordination is restored through a shared latent embedding.
    \item \textbf{Foundation-model-guided coordination.} A frozen CLIP model with some trainable projectors encodes global RGB views, robot states, and language instructions into a latent representation that conditions both diffusion policies. 
    This foundation-model-guided latent serves as a lightweight, semantically rich consensus channel, distinguishing our approach from end-to-end VLA controllers and from strict learned consensus mechanisms such as LatentToM.
    \item \textbf{Phase-aware semantic coordination with contrastive alignment.} We introduce a phase-progress head that uses a frozen CLIP model to encode multi-view RGB observations, proprioceptive states, task-level language instruction, and hand-crafted textual descriptions of task phases into a shared latent space. 
    From these text prompts, the model infers both discrete phase scores and continuous progress estimates without any annotated phase labels or additional supervised heads. 
    Meanwhile, we impose a coordination-aware contrastive loss that pulls the latent towards correctly paired arm-base joint actions and pushes it away from mismatched arm/base combinations, explicitly shaping the embedding to capture cross-subsystem compatibility.
\end{itemize}
We evaluate FALCON on two challenging loco-manipulation tasks that require long-horizon navigation, precise end-effector placement, and tight coupling between base motion and manipulator behavior. 
Across these tasks, FALCON outperforms the centralized and decentralized baselines, while exhibiting improved robustness and generalization to out-of-distribution (OOD) scenarios.
These OOD scenarios may arise from observation mismatches between the manipulator and the quadruped, or from the inherent dynamics and instability of legged locomotion. 
Our results suggest that, in the emerging landscape of robot learning and foundation models, combining decoupled RGB-based visuomotor specialists with foundation-model-guided coordination offers a promising alternative to fully monolithic whole-body policies.

\section{Related Works}

\subsection{Mobile Manipulation}

Mobile manipulation integrates mobility and dexterous interaction within a single platform, enabling robots to act on objects across large, cluttered workspaces. This capability is central to many long-horizon tasks and has been studied across diverse morphologies, from wheeled mobile manipulators to humanoids and legged systems~\cite{billard2019trends, bai2025towards, wang2025robot,gong2023legged}. Recent surveys emphasize that combining locomotion and manipulation stresses planning, control, and perception stacks simultaneously, especially when interaction must be robust in unstructured environments~\cite{bai2025towards, wang2025robot}. Within this landscape, current trends revolve around base morphology, base-arm coordination, control scheme and perception.

Classical wheeled mobile manipulators focused on navigation-plus-grasp pipelines: a mobile base moves to a suitable pose, then a fixed-base arm executes a manipulation plan~\cite{chitta2012mobile, sandakalum2022motion}. Representative systems demonstrate pick-and-place and door-opening in unstructured indoor environments through integrated perception, task planning, and motion planning~\cite{chitta2012mobile}. Recent work increasingly replaces hand-engineered stacks with learning-based controllers. 
For example, N²M² uses a wheeled mobile manipulator to perform long-horizon mobile manipulation tasks in unstructured environments, relying on learned navigation policies that generate feasible base motions for arbitrary end-effector trajectories while handling dynamic obstacles~\cite{honerkamp2023n}. 
ReLMM further shows that fully autonomous reinforcement learning can continuously improve real-world mobile manipulation performance through extended autonomous interaction~\cite{sun2022fully}.
Task-focused systems such as TidyBot exploit language-conditioned policies on a wheeled base to declutter a user’s room according to personalized preferences~\cite{wu2023tidybot}. Large-scale imitation and RL pipelines (RT-1/RT-2) demonstrate that end-to-end visuomotor learning from extensive real-world data can yield robust pick-and-place and tool-use behaviors on mobile platforms using only onboard cameras and minimal task engineering~\cite{brohan2022rt, zitkovich2023rt}. These works collectively show that learning-based controllers can absorb the complexity of real environments more flexibly than purely model-based stacks, but most still treat base and arm as a single centralized policy and rely on relatively slow receding-horizon planners.

In parallel, legged robots have progressed from pure locomotion toward loco-manipulation, with early works using optimization or hybrid optimization–learning methods already demonstrating the importance of whole-body coordination in manipulation~\cite{zimmermann2021go, chiu2022collision, ma2022combining}. More recently, the control paradigm has shifted toward learning-based approaches, which can capture complex whole-body dynamics without extensive hand-engineered modeling~\cite{fu2023deep}. In addition, learning-based methods unlock more advanced capabilities such as force-aware control~\cite{portela2024learning, zhi2025learning}, multi-contact planning~\cite{sleiman2023versatile}, body motion modulation~\cite{gu2025humanoid}, and inter-limb coordination~\cite{zhu2025versatile, yang2025helom}. The emergence of recent systems such as UMI-on-Legs~\cite{ha2024umi} further advances loco-manipulation autonomy by using a behavior-clone diffusion policy as high-level guidance, replacing traditional teleoperation inputs. Analyses of multi-legged manipulators further support the view that distributing load-bearing and task-executing roles across different legs or limbs yields better stability and dexterity in cluttered terrain~\cite{gong2023legged}. These developments motivate using a legged platform not only for mobility but also to actively modulate the arm’s pose by bending, crouching, or re-centering the body, effectively turning whole-body motion into an additional degree of freedom for manipulation.

Modern manipulation systems increasingly rely on high-capacity sequence models that treat control as conditional sequence generation~\cite{shridhar2023perceiver, shafiullah2022behavior}. Transformer-based policies (Perceiver-Actor, Behavior Transformers, RT-1) learn multi-task manipulation skills from mixed demonstration datasets by modeling multi-modal action distributions over long horizons~\cite{shridhar2023perceiver, shafiullah2022behavior, brohan2022rt}. Diffusion policies further refine this idea by modeling action trajectories as samples from a learned denoising process, providing strong multi-modality, stability, and sample efficiency in contact-rich visuomotor tasks~\cite{chi2025diffusion, liu2025diffusion, wolf2025diffusion}. Diffusion-based controllers have been applied to a range of manipulation problems, and recent work extends them to mobile settings: GNFactor uses diffusion-like latent representations for multi-task real-robot learning~\cite{ze2023gnfactor}, while M4Diffuser explores multi-view diffusion policies and manipulability-aware control for robust mobile manipulation~\cite{dong2025m4diffuser}. Diffusion surveys emphasize that such models naturally support stochastic action sampling, uncertainty-aware rollouts, and flexible conditioning on auxiliary signals (e.g., goals or language)~\cite{liu2025diffusion, wolf2025diffusion}. In loco-manipulation, these properties are particularly appealing for robust decentralized coordination, where a locomotion policy and a manipulation policy each solve a complex subproblem yet must remain consistent through a shared latent context. Decentralized diffusion architectures for cooperative manipulation, such as LatentToM, demonstrate that decoupled agents can achieve high-quality decision making while coordinating via shared semantic latent variables rather than explicit centralized control~\cite{he2025latent}. This provides a natural template for splitting base and arm control on a mobile (e.g. quadruped) platform while still achieving coherent whole-body behavior.

Across these lines of work, there is a clear trend toward end-to-end visuomotor learning that directly maps images to actions~\cite{wang2025robot, bai2025towards, shafiullah2022behavior}. Systems such as CLIPort and Perceiver-Actor show that RGB-only inputs, when combined with strong visual encoders and task conditioning, can support precise pick-and-place and tool-use in cluttered scenes~\cite{shridhar2022cliport, shridhar2023perceiver}. Large-scale grasping and mobile manipulation frameworks (QT-Opt, BC-Z, RT-1) likewise achieve robust performance using monocular or multi-view RGB cameras without requiring active depth sensing~\cite{dmitry2018qt, jang2022bc, brohan2022rt}. Surveys on embodied visual perception argue that dense, overlapping RGB views can provide sufficient geometric cues for many manipulation tasks, especially when combined with temporal information and learned feature representations~\cite{wang2025robot}. In mobile manipulation, the placement of cameras is an important design choice, and using multiple views from different locations on the robot provides both global context and local detail for control. Such multi-camera configurations, as adopted in recent mobile and bimanual manipulation systems~\cite{sun2022fully, fu2024mobile, brohan2022rt}, align with the broader shift toward camera-only, learning-based perception pipelines that trade hardware complexity for algorithmic inference.

\begin{table*}[t]
\centering
\caption{Related characteristics of recent manipulation works}
\label{tab:major_works_comparison}
\begin{tabular}{@{}ccccccc@{}}
\toprule
Robot/Project           & Year & Foundation Model & Learning-based      & Visual Modality & Legged-based & Decentralized \\ \midrule
CLIPort~\cite{shridhar2022cliport}                 & 2021 & \Checkmark              & \Checkmark                 & RGB-D           & \XSolidBrush                & \XSolidBrush            \\
Go Fetch!~\cite{zimmermann2021go}               & 2021 & \XSolidBrush               & \XSolidBrush                  & RGB             & \Checkmark               & \XSolidBrush            \\
ALMA~\cite{ma2022combining}                    & 2022 & \XSolidBrush               & \Checkmark(body) + \XSolidBrush(arm) & -             & \Checkmark               & \Checkmark           \\
Collision-Free MPC~\cite{chiu2022collision}      & 2022 & \XSolidBrush               & \XSolidBrush                  & -           & \Checkmark               & \XSolidBrush            \\
Deep Whole-Body Control~\cite{fu2023deep} & 2022 & \XSolidBrush               & \Checkmark                 & RGB           & \Checkmark               & \XSolidBrush            \\
SayCan~\cite{ahn2022can}                  & 2022 & \Checkmark              & \Checkmark                 & RGB             & \XSolidBrush               & \XSolidBrush            \\
ACT~\cite{fu2024mobile}            & 2024 & \XSolidBrush               & \Checkmark                 & RGB             & \XSolidBrush               & \XSolidBrush            \\
$\pi_0$~\cite{black2024pi0}                     & 2024 & \Checkmark              & \Checkmark                 & RGB             & \XSolidBrush               & \XSolidBrush            \\
UMI on Legs~\cite{ha2024umi}                     & 2024 & \Checkmark              & \Checkmark                 & RGB             & \Checkmark               & \XSolidBrush            \\
ELLMER~\cite{mon2025embodied}                  & 2025 & \Checkmark              & \Checkmark                 & RGB-D             & \XSolidBrush                & \XSolidBrush            \\
GR00T N1~\cite{bjorck2025gr00t}/Dream~\cite{jang2025dreamgen}          & 2025 & \Checkmark              & \Checkmark                 & RGB             & \XSolidBrush                & \XSolidBrush            \\
LatentToM~\cite{he2025latent}               & 2025 & \XSolidBrush               & \Checkmark                 & RGB             & \XSolidBrush                & \Checkmark           \\
ReLIC~\cite{zhu2025versatile}                   & 2025 & \Checkmark              & \Checkmark                 & RGB-D             & \Checkmark               & \XSolidBrush            \\
UniFP~\cite{zhi2025learning}                   & 2025 & \XSolidBrush               & \Checkmark                 & RGB           & \Checkmark               & \XSolidBrush            \\
\textbf{FALCON(ours)}               & - & \Checkmark             & \Checkmark                & RGB             & \Checkmark               & \Checkmark            \\\bottomrule
\end{tabular}
\end{table*}

\subsection{Foundation Model for Robotics}

Foundation models have recently emerged as a promising solution to the key limitations of traditional learning-based robotics. Classical approaches typically learn task-specific policies from scratch. This specialization incurs huge data demands. For example, recent work notes that these policies often require hundreds of thousands of real-robot trials to master even basic tasks~\cite{ho2021sim2real}. Even with massive data, such policies tend to overfit narrow domains~\cite{khan2025foundation,meng2025preserving,he2025demystifying}. In short, classical approaches struggle with generalization, transfer, and efficiency~\cite{firoozi2025foundation,black2024pi0}. These challenges motivate leveraging foundation models as a new paradigm in robotics. By importing broad semantic priors and multi-modal understanding, foundation models offer a path to more flexible, general-purpose robot behavior~\cite{khan2025foundation}.

Typically, pretrained foundation models offer substantial advantages by learning semantic representations and cross-modal alignments from large-scale, diverse data sources, which can be transferred to downstream robotics tasks.
For instance, large language models such as GPT exhibit strong commonsense reasoning and even code generation capabilities without task-specific fine-tuning, while vision-language models like CLIP support open-vocabulary visual recognition and semantic retrieval~\cite{khan2025foundation}. 
Foundation models significantly enhance data efficiency and generalization. Robots can learn new tasks from limited demonstrations and generalize to novel environments in a zero-shot manner.
In addition, foundation models can serve as unified representation backbones throughout the perception–planning–control pipeline, providing a shared embedding space for vision, language, and action. This consolidation reduces the need for task-specific components and supports more extensible and general-purpose robotic systems~\cite{zeng2023large}. In what follows, the applications of foundation models in robotics across four major dimensions are discussed.

Foundation models have become strong open-vocabulary visual backbones for robotic perception and scene understanding. Vision-language models like CLIP furnish joint image–text embeddings that allow robots to recognize arbitrary objects without task-specific retraining~\cite{radford2021learning}, while SAM provides prompt-driven, category-agnostic segmentation that functions as an open-set detector~\cite{kirillov2023segment}.
Temporal encoders such as R3M, pretrained on large-scale video corpora like Ego4D, supply spatio-temporal features that improve manipulation in dynamic settings. Studies report over 20\% gains compared to training encoders from scratch and it enables a real robot to learn new skills from only 20 demonstrations~\cite{nair2022r3m,grauman2022ego4d}.
Complementing these models, open-vocabulary detectors such as Detic, OWL-ViT, and ViLD leverage CLIP-style vision–language representations to enable zero-shot detection of novel object categories~\cite{zhou2022detecting,minderer2022simple,gu2021open}.
Foundation models also enable semantic mapping. LM-Nav leverages CLIP embeddings to associate observations with language-grounded landmarks~\cite{shah2023lm}, while CLIPFields integrates CLIP features into 3D reconstructions, achieving semantically meaningful 3D maps~\cite{shafiullah2022clip}.

At a higher level, large pre-trained language and vision–language foundation models have begun to handle task planning and instruction understanding. Approaches such as SayCan couple LLM-based reasoning with robot affordances: given a natural-language goal, the LLM proposes candidate action steps and ranks them by both semantic relevance and physical feasibility. Specifically, SayCan uses a PaLM-based LLM to decompose a high-level instruction into subtasks, and then applies learned affordance value functions to filter out infeasible actions. This produces interpretable, closed-loop task plans grounded in the robot’s capabilities~\cite{ahn2022can,chowdhery2023palm}.
Likewise, PaLM-E interleaves multimodal input to generate sequential plans. Experiments show that PaLM-E, as a single large model, can perform diverse embodied tasks ranging from tabletop manipulation to vision-based question answering, benefiting from positive transfer across modalities~\cite{driess2023palm}. More broadly, these foundation models can parse complex instructions, employ chain-of-thought reasoning to infer intermediate steps, and adapt task plans on the fly, enabling robots to execute language-specified workflows with greater flexibility~\cite{wei2022chain,brohan2022rt}.

Recent advances in foundation models have enabled manipulation policies that map visual observations and natural language instructions to low-level actions. Early approaches such as CLIPort combine CLIP’s pretrained semantic representations with spatial pathway to execute a variety of language-specified pick-and-place tasks in both simulation and the real world. These models exhibit strong data efficiency and zero-shot generalization to unseen object-instruction combinations~\cite{shridhar2022cliport}. 
PerAct further extends this idea to 6-DoF control by encoding RGB-D inputs and goal texts with a Perceiver Transformer, achieving robust multi-task manipulation with few demonstrations~\cite{shridhar2023perceiver}.
More recent systems like RT-2 and OpenVLA unify Internet-scale vision-language pretraining with robotic trajectories, forming scalable vision-language-action models that generalize across skills, objects, and embodiments~\cite{zitkovich2023rt,kim2024openvla,o2024open}. 
Collectively, these approaches demonstrate that vision-language-conditioned policies provide a practical and flexible interface for real-world manipulation, enabling robots to follow diverse user instructions and adapt to new tasks with minimal additional training.

A major trajectory in robotics is the development of unified multimodal models that integrate perception, language, and control into a single system. These models aim to generalize across tasks, environments, and robot embodiments. 
A pioneering effort in this direction is Gato, which is a 1.18B-parameter transformer trained on a large mixture of datasets. 
It ingests multi-modality inputs and produces either text or action tokens. 
With fixed weights, the same policy can play Atari games, caption images, engage in dialogue, and manipulate blocks with a real robot arm. This illustrates that a single foundation model can span language, vision, and control~\cite{reed2022generalist}. 
PaLM-E similarly integrates visual, proprioceptive, and linguistic inputs into a unified embodied large model for high-level reasoning and planning in robotic tasks~\cite{driess2023palm}.
$\pi_0$ demonstrated a generalist robot policy built on a frozen vision–language backbone and a diffusion-flow action model. Trained on a rich multi-embodiment dataset spanning single-arm, dual-arm, and mobile manipulators, $\pi_0$ can be prompted or fine-tuned to perform new tasks. On real robots, it folds laundry, cleans tables, and assembles boxes using the same underlying representations~\cite{black2024pi0}. ELLMER enables grounded multi-step reasoning for embodied control, and GR00T-Dreams leverages language-conditioned video world models to generate large-scale synthetic trajectories that improve policy generalization~\cite{mon2025embodied,jang2025dreamgen}. These results suggest that truly generalist, multimodal robot agents are increasingly within reach through foundation-model techniques.

\section{Problem Formulation}

This work focuses on the loco-manipulation problem for legged robots equipped with a mounted manipulator. The robot must navigate through the environment, position its base appropriately, and execute precise end-effector interactions to complete long-horizon tasks such as opening a drawer or placing an object. Loco-manipulation is inherently multimodal: the locomotion subsystem relies primarily on proprioception and base motion dynamics, while the manipulation subsystem depends on fine-grained geometric cues around the target object. These sensing and control modalities differ significantly in scale, bandwidth, and timescales.

A loco-manipulation task is defined as a sequence of navigation and interaction phases that the robot must execute coherently. The robot must (1) travel from its initial pose to a manipulation-capable region, (2) achieve an appropriate whole-body configuration that places the end-effector in a useful workspace, and (3) perform precise manipulation actions to accomplish the task objective.

Conventional whole-body control methods attempt to fuse these heterogeneous observations into a single policy, which often leads to instability, conflicting gradients, and degraded performance, especially when the robot must operate under visual ambiguity or distribution shift. In contrast, our framework adopts a decoupled formulation, where locomotion and manipulation are treated as two independent visuomotor subsystems, each operating in its own natural observation space and action space. The locomotion subsystem produces high-level velocity and pose commands that are executed by a low-level reinforcement-learning-based controller, while the manipulation subsystem produces high-level end-effector targets executed through inverse kinematics. 

While decoupling enables each subsystem to specialize, it introduces a new fundamental challenge: restoring coordination between subsystems that no longer share observations or action spaces. The core objective of the problem is therefore to enable coordinated, task-aware decision making between two independent policies, such that the locomotion subsystem positions and orients the base in ways that are compatible with the manipulator’s needs, and the manipulation subsystem adapts its actions to the motion and pose of the base.

Our method seeks to learn a unified decision-making structure that aligns these decoupled policies so they can jointly complete complex loco-manipulation tasks, while preserving the modularity, robustness, and generalization benefits afforded by the decoupled design.

\section{Decoupled Loco-Manipulation}

\begin{figure*}
    \centering
    \includegraphics[width=1\linewidth]{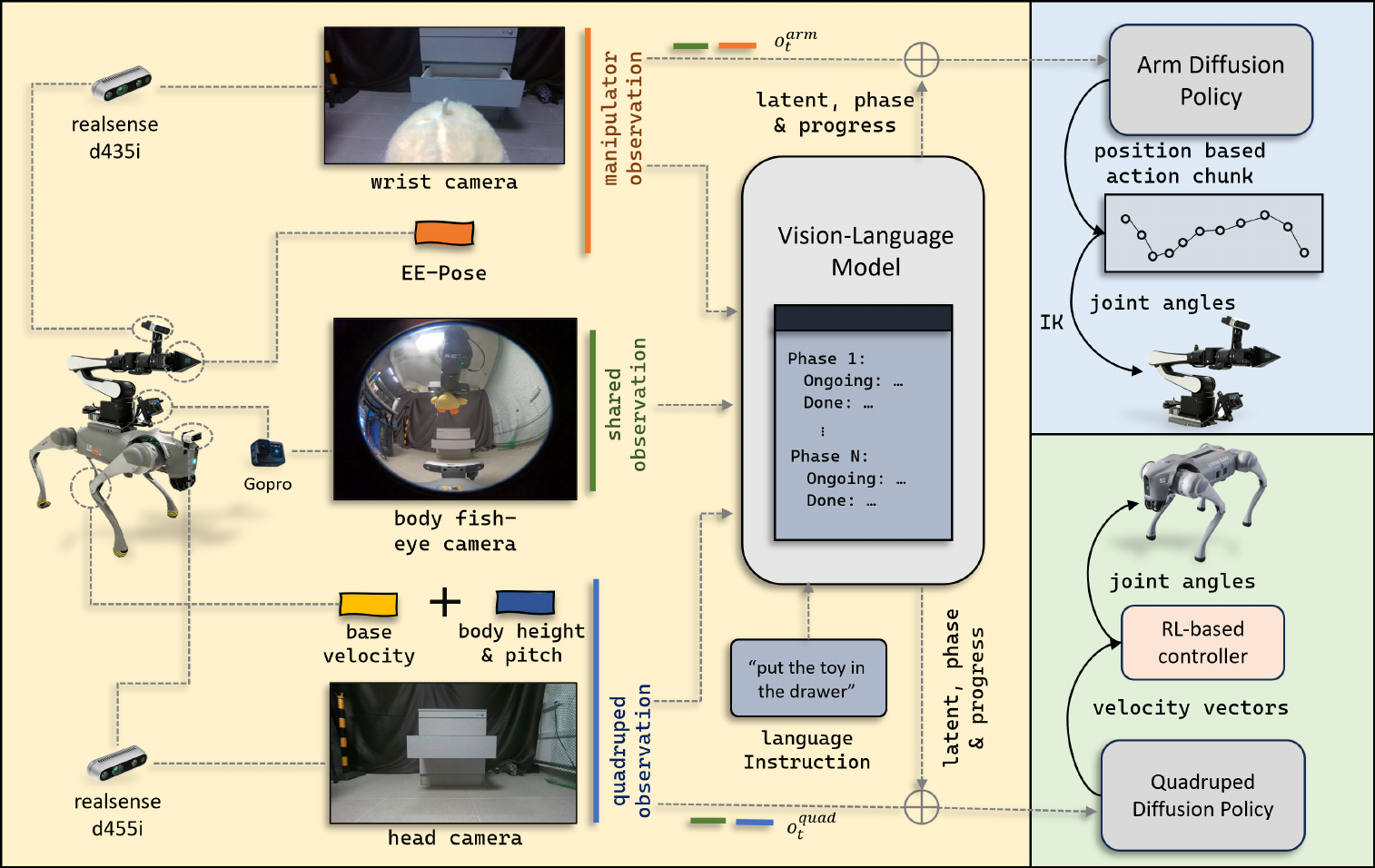}
    \caption{Overall framework of FALCON. The blue and green regions denote the decoupled manipulator and quadruped diffusion policies, which act in their own observation and control spaces (wrist/body cameras and end-effector pose for the arm; head/body cameras, base velocity, and body height/pitch for the quadruped). The yellow region shows the foundation model module, which aggregates global RGB observations and the language instruction into a shared latent, while capture task phase and progress simultaneously; this task-representative latent is then used to jointly condition both diffusion policies, providing semantic coordination between the two subsystems.}
    \label{fig:overall_framework}
\end{figure*}

\subsection{Core Insight}

The central insight of this work is that, within loco-manipulation tasks, decoupled policies enable each subsystem to achieve higher-quality decision making (especially for quadruped policy according to our experiments), while a foundation model provides a shared semantic context that facilitates coordination between the two otherwise independent control streams.
Formally, we define two diffusion policies, one for locomotion and one for manipulation, and a shared latent vector generated by a vision-language foundation model (CLIP) as
\begin{equation}\label{eq:iv.a1}
    \begin{aligned}
    a_t^{quad}\sim\pi_\theta^{quad}(o_t^{quad},z_t),~~~a_t^{arm}\sim\pi_\phi^{arm}(o_t^{arm},z_t),
    \end{aligned}
\end{equation}
where $o_t^{quad}$ and $o_t^{arm}$ denote the local observation of the locomotion and manipulation subsystems, respectively.
$\pi_{\theta/\phi}^{quad/arm}$ and $a_t^{quad/arm}$ represent the diffusion policy and action of the quadruped/manipulator. 
For the quadruped, its action includes velocity and pose; and for the manipulator, its action is the end-effector pose.
The shared latent embedding $z_t$ encodes contextual information from the foundation model:
\begin{equation}\label{eq:iv.a2}
    \begin{aligned}
        z_t = f_{CLIP}(I_t,s_t,p_{text})
    \end{aligned}
\end{equation}
where $I_t$ here represents the whole system's RGB visual observations from the environment, $s_t$ denotes proprioceptive/global robot states, and $p_{text}$ here is a text-prompt which contains the overall task's context.
This latent vector $z_t$ is concatenated with each subsystem’s observation to form the conditioning input of the diffusion models, enabling both modules to operate decoupled yet remain semantically aligned.

As illustrated in Fig~\ref{fig:overall_framework}, a pre-trained reinforcement learning low-level controller $\pi_\psi^{LLC}$ governs the low-level joint-angle control of the quadruped:
\begin{equation}\label{eq:iv.a3}
    \begin{aligned}
    a_t^{LLC}=\pi_\psi^{LLC}(s_t^{LLC}),
    \end{aligned}
\end{equation}
where $a_t^{LLC}$ is motor-level command and $s_t^{LLC}$ is current quadruped's locomotion state.
On top of this low-level controller, a diffusion policy produces high-level velocity and pose commands:
\begin{equation}\label{eq:iv.a4}
    \begin{aligned}
        a_t^{\text{quad}}\coloneqq s_t^{quad}=\pi_\theta^{quad}(o_t^{quad},z_t)
    \end{aligned}
\end{equation}
which are executed in an open-loop manner.
The output of diffusion policy $a_t^{\text{quad}}\coloneqq s_t^{quad}$ works as part of the input of $\pi_\psi^{LLC}$. Details can be found in Eq~\ref{eq:s_llc}.
We adopt velocity control instead of position control because our visuomotor policy relies solely on RGB inputs, without access to depth or point-cloud data typically required for precise position-based navigation.
This open-loop velocity regime imposes additional challenges in stability and error accumulation, as analyzed in Section~\ref{sec:intro}.
For the manipulator mounted atop the quadruped, we employ a separate diffusion policy that outputs position-based high-level commands:
\begin{equation}\label{eq:iv.a5}
    \begin{aligned}
        a_t^{arm}\coloneqq x_t^{arm}=\pi_\phi^{arm}(o_t^{arm},z_t),
    \end{aligned}
\end{equation}
which are converted to joint targets through an inverse-kinematics (IK) solver $u_t^{arm}=\mathtt{IK}(x_t^{arm})$.
Unlike locomotion, the manipulator operates under position control, which provides higher accuracy and better stability.
Decoupling locomotion and manipulation into two independent policies thus allows each subsystem to maintain robustness under its own control regime, while the shared latent embedding $z_t$ ensures coordinated behavior.

To achieve semantic and behavioral consistency across the two decoupled policies, we incorporate a vision-language foundation model (CLIP) as a context synthesizer.
CLIP processes multi-modal observations, RGB images, robots states, and language cues, to produce the latent vector $z_t$ in Eq~(\ref{eq:iv.a2}).
This latent embedding captures task-relevant semantics and environmental context, serving as a shared prior that aligns the action generation of both subsystems:
\begin{equation}
    \begin{aligned}
        \pi^{quad},\pi^{arm}\Leftarrow\mathtt{Cond}~z_t=f_{CLIP}(I_t,s_t,p_{text})
    \end{aligned}
\end{equation}
Consequently, locomotion and manipulation remain physically decoupled but semantically coupled through the latent representation, enabling coherent high-level decision making and synchronized action generation across the entire loco-manipulation system.

\subsection{Loco-Manipulation System Setup}

\label{subsec:loco_manipulation_system_setup}
We consider a quadruped mobile manipulator equipped with a legged base and a mounted arm. Our goal is to enable loco-manipulation in indoor environments, where the robot must maintain both a wide field of view for global information and precise visual feedback around the end effector.

To support robust maneuvering, we employ a tri-camera configuration with complementary perspectives. A wrist-mounted camera provides close-range observations of the end effector and nearby objects. For example, in the pick-and-place task, it allows the system to verify that an object is securely grasped by the gripper and that it is placed accurately at the target location.
A body-mounted camera provides a wide-angle egocentric view that captures the overall arm configuration and the base motion during mobile manipulation. Additionally, a head-mounted camera offers a lower-angle perspective of the surrounding workspace, which is used to visually guide the base when approaching the target area.

All cameras operate in RGB-only mode, without using depth measurements. Relying on appearance cues simplifies the perception pipeline, while the multi-view setup compensates for occlusions and viewpoint-specific ambiguities. Overall, this design provides the control policy with rich visual information at different spatial scales, which is essential for performing reliable maneuvers on real robotic systems.

\subsection{Low-level Locomotion Controller}

To achieve effective decoupling between the legged base and the mounted manipulator, we employ a learning-based low-level locomotion controller. This controller is trained entirely in simulation and subsequently deployed on the physical robot to function as the mobile base controller. It is optimized via reinforcement learning to track high-level velocity and pose commands while simultaneously regulating body orientation and height, thereby ensuring stable, dynamically consistent, and robust whole-body motion.

The training of the low-level controller is formulated as a Markov Decision Process (MDP),
\begin{equation}
\mathcal{M} = (\mathcal{S}, \mathcal{A}, \mathcal{P}, r, \gamma),
\end{equation}
where $\mathcal{S}$ and $\mathcal{A}$ denote the state and action spaces, $\mathcal{P}$ the transition dynamics, $r$ the reward function, and $\gamma \in (0,1)$ the discount factor.

At time step $t$, the system state $s_t \in \mathcal{S}$ aggregates proprioceptive measurements together with high-level command signals:
\begin{equation}\label{eq:s_llc}
s_t^{\text{LLC}} =
\big[
q_t,\;
\dot{q}_t,\;
g_t,\;
\omega_t,\;
a_t^{\text{quad}}
\big],
\end{equation}
where $q_t$ and $\dot{q}_t$ denote the joint positions and velocities, $g_t$ is the gravity vector expressed in the robot frame, and $\omega_t$ is the base angular velocity.  
The high-level command vector is defined as
\begin{equation}
a_t^{\text{quad}}
=
\big[
\mathbf{v}^{\text{cmd}}_t,\;
\theta^{\text{cmd}}_t,\;
h^{\text{cmd}}_t
\big],
\end{equation}
where $\mathbf{v}^{\text{cmd}}_t$ is the commanded base linear and angular velocity, $\theta^{\text{cmd}}_t$ is the desired body pitch angle, and $h^{\text{cmd}}_t$ denotes the target base height.
The action $u_t^{joint}$ represents the joint-level command generated by the low-level locomotion policy $\pi_\psi^{LLC}$ mentioned in Eq.\ref{eq:iv.a3}.

We adopt a relatively simple reward formulation for the blind walking scenario, consistent with prior research~\cite{long2023hybrid}, and augment it with additional task-specific rewards for base height and body orientation tracking to better support stable whole-body manipulation.

To enforce torso orientation stability, an orientation control reward is introduced to penalize deviations from the commanded roll and pitch angles. Let
$\boldsymbol{\theta}_t^{\text{cmd}} = [\theta_t^{\text{roll}}, \theta_t^{\text{pitch}}]$ denote the commanded body orientation, and let $\mathbf{g}_t \in \mathbb{R}^3$ represent the gravity vector expressed in the robot base frame. The desired gravity vector $\mathbf{g}_t^{\text{des}}$, corresponding to the target orientation, is obtained by rotating the nominal gravity vector from the world frame into the base frame using the inverse of the desired base orientation:

\begin{equation}
\mathbf{g}_t^{\text{des}}
= \mathbf{R}^{\top}\big(\boldsymbol{q}_t^{\text{des}}\big) \mathbf{g}_0,
\end{equation}
where $\boldsymbol{q}_t^{\text{des}}$ is the quaternion constructed from $\boldsymbol{\theta}_t^{\text{cmd}}$, $\mathbf{R}(\cdot)$ denotes the rotation matrix associated with the corresponding quaternion, and $\mathbf{g}_0$ is the nominal gravity vector expressed in the world frame.
The orientation tracking reward is then defined as
\begin{equation}
r_t^{\text{ori}} =
\left|
\mathbf{g}_t^{xy} - \mathbf{g}_t^{\text{des},xy}
\right|_2^2,
\end{equation}
where $(\cdot)^{xy}$ denotes the projection of a vector onto the horizontal plane. This formulation encourages the base to maintain the desired roll and pitch angles by minimizing discrepancies between the measured and desired projected gravity directions.

The policy $\pi_\psi^{LLC}(a_t^{LLC}\mid s_t^{\text{LLC}})$ is optimized using the Proximal Policy Optimization (PPO)~\cite{schulman2017proximal} algorithm to maximize the expected discounted return:
\begin{equation}
J(\theta)
=
\mathbb{E}_{\pi_\psi^{LLC}}
\left[
\sum_{t=0}^{\infty} \gamma^t r_t
\right].
\end{equation}
where $r_t$ denotes the immediate reward received at time $t$ after executing action $a_t^{LLC}$ in state $s_t^{\text{LLC}}$.

To enhance sim-to-real transfer and operational robustness, domain randomization is applied during training, incorporating variations in terrain profiles, friction coefficients, payload configurations, and actuator dynamics. Once trained, the controller provides a reliable and responsive mobile base that shields higher-level modules from the complexity of legged dynamics. This enables the quadruped to be treated as a velocity- and pose-controllable platform, allowing the high-level diffusion policy to operate purely in command space without direct interaction with low-level joint dynamics.

\subsection{Foundation Model Informed Coordination}

The core insight of this work is that, within loco-manipulation tasks, explicitly decoupling the control policies for locomotion and manipulation, rather than relying on a monolithic whole-body controller, allows each subsystem to achieve more specialized and higher-quality decision making. 
However, such decoupling inherently introduces a new challenge: \textit{how to ensure effective cooperation and temporal consistency between two independently optimized policies?}
Recent advances in decentralized diffusion architectures have shown that coordination can emerge through consensus embeddings formed during training~\cite{he2025latent}, enabling decentralized manipulators to share a unified representation of the task.
Inspired by this paradigm, our framework adopts a conceptually similar yet practically distinct approach. 
Instead of pursuing strict decentralization across separate computational agents, we deliberately decouple the loco-manipulation system into two specialized sub-policies to enhance modularity and robustness, while maintaining a shared semantic channel for coordination.

To achieve this coordination, we leverage a vision foundation model, specifically CLIP, a large-scale vision-language model pre-trained on Internet-scale data. 
As illustrated in Fig~\ref{fig:overall_framework}, CLIP processes visual observations and high-level task cues to generate a latent embedding that serves as a common contextual representation, aligning the independent locomotion and manipulation policies toward coherent, goal-directed behavior.

\subsubsection{Contrastive loss for temporal consistency}

Due to CLIP being not specifically trained for robot related tasks, we require an adapter to fine-tune the raw output by our dataset to generate the desired $z_t$.
For each training sample $i\in\{ 1,\cdots,B\}$, we have a short sequence of observations and a joint action chunk for the loco-manipulation system. 
The VLM backbone produces per-frame latent features
\begin{equation}
    \begin{aligned}
        z_{i,t}\in\mathbb{R}^D,~~t=1,\cdots,T_{obs}
    \end{aligned}
\end{equation}
which we aggregate into a single per-sample latent by temporal averaging:
\begin{equation}
    \begin{aligned}
        \bar{z}_i=\frac{1}{T_{obs}}\sum^{T_{obs}}_{t=1}z_{i,t}\in\mathbb{R}^D
    \end{aligned}
\end{equation}
For the actions, the manipulator and quadruped actions are defined as:
\begin{equation}
    \begin{aligned}
        a_{i,h}^{arm}\coloneq x_{t+h-1},~~a_{i,h}^{quad}\coloneq v_{t+h-1},~h=1,\cdots,H
    \end{aligned}
\end{equation}
We summarize each horizon by averaging over time:
\begin{equation}
    \begin{aligned}
        \bar{a}_i^{arm/quad}=\frac{1}{H}\sum_{h=1}^H a_{i,h}^{arm/quad}
    \end{aligned}
\end{equation}
and form a joint action for sample $i$ by concatenating them $u_i=[\bar{a}_i^{arm};\bar{a}_i^{quad}]$.
To map observation and joint action summaries into a common contrastive space, we use two MLP projection heads to map both $\bar{z}_{i,t}$ and $u_i$ to the same space and become $v_i$ and $w_i$.
To explicitly encode two subsystems' coordination, we construct mismatched joint action negative pairs by shuffling manipulator and quadruped summaries across samples.
Let $\lambda$ be a random permutation of $\{ 1,\cdots,B\}$ with no fixed points $\lambda(i)\neq i$.
Therefore, there will be two types of mismatched joint actions:
\begin{equation}
    \begin{aligned}
        u_i^{(arm)}=[\bar{a}_i^{arm};\bar{a}_{\lambda(i)}^{quad}],
       ~u_i^{(quad)}=[\bar{a}_{\lambda(i)}^{arm};\bar{a}_i^{quad}].
    \end{aligned}
\end{equation}
Similarly, we then encode them as $w_i^{(arm)}$ and $w_i^{(quad)}$.
With a temperature $\tau>0$, we define an InfoNCE-style contrastive loss~\cite{oord2018representation} that pulls each observation embedding $v_i$ towards its coordinated joint action $w_i$ and pushes it away from all other candidates:
\begin{equation}
    \begin{aligned}
        \mathcal{L}_{z\rightarrow u} = - \frac{1}{B}\sum_{i=1}^B\log\frac{\exp (\frac{1}{\tau}v_i^Tw_i)}{\sum_{c\in\mathcal{C}}\exp(\frac{1}{\tau}v_i^Tc)}.
    \end{aligned}
\end{equation}
To make sure the coordinated action also can match its corresponding observation, we have another similar loss which can be expressed as $\mathcal{L}_{u\rightarrow z}$.
The final coordination-aware contrastive loss is:
\begin{equation}
    \begin{aligned}
        \mathcal{L}_{coord}=\frac{1}{2}(\mathcal{L}_{z\rightarrow u}+\mathcal{L}_{u\rightarrow z})
    \end{aligned}
\end{equation}
and is combined with the diffusion-policy losses for the manipulator and quadruped controllers as
\begin{equation}
    \begin{aligned}
        \mathcal{L}_{tot}=\mathcal{L}_{diff}^{(arm)}+\mathcal{L}_{diff}^{(quad)}+\delta \mathcal{L}_{coord}
    \end{aligned}
\end{equation}
where $\delta$ weights the strength of the coordination regularizer. $\mathcal{L}_{diff}^{(arm)}$ and $\mathcal{L}_{diff}^{(quad)}$ represent the diffusion policy loss of the manipulator and quadruped, respectively.

We use this contrastive loss as an explicit coordination regularizer on the VLM latent. 
Intuitively, each observation embedding is encouraged to align with the coordinated joint action that was actually executed under that observation, while being repelled from ``mismatched'' joint actions where the manipulator and quadruped components are taken from different samples. 
By treating these mismatched combinations as hard negatives in the contrastive loss, the latent representation is driven to encode not only what each subsystem should do individually, but also how their actions must be mutually consistent to achieve coherent whole-body loco-manipulation.

\subsubsection{Language-defined phase and progress head} 

On top of a frozen CLIP backbone, we first build a global visual–proprioceptive embedding. Given multi-view RGB observations and proprioceptive states at time $t$, we compute:
\begin{itemize}
    \item CLIP image features for each camera view,
    \item an output of MLP encoder over proprioception (end-effector pose and robot base state),
    \item and a CLIP text embedding of the language instruction.
\end{itemize}
These tokens are fused with a shallow Transformer encoder, yielding a global embedding $h_t\in\mathbb{R}^{512}$, which already captures multi-view visual context, robot state, and the high-level goal.
To make the task structure explicit, we define a small set of human-interpretable phases ${\phi_k}_{k=1}^K$ in natural language. 
In our Navigation and Placement task, we set four phases as \texttt{move}, \texttt{align}, \texttt{place}, and \texttt{close}~\cite{huang2024rekep,Krishnan2017LfDSurgical}.
Each phase $\phi_k$ is specified by an \texttt{ongoing} and a \texttt{done} prompt, e.g.
\emph{the robot is closing the drawer} vs. \emph{the drawer is fully closed}.
We pre-encode these prompts with the CLIP text encoder, obtaining embeddings $e^{\text{ongoing}}_k$ and $e^{\text{done}}_k$.

At each time step $t$, we compare the current image embedding $f_t$ with the phase prompts to obtain phase logits and probabilities
\begin{equation}
    \begin{aligned}
        \ell_{t,k}=\langle f_t,e_k^{ongoing}\rangle,~~\rho_t=\text{softmax}(\ell_t)\in\mathbb{R}^K, 
    \end{aligned}
\end{equation}
where $\rho_{t,k}$ can be interpreted as a soft belief that the system is currently in phase $\phi_k$.
To capture how far the task has progressed within each phase~\cite{kang2025incorporating}, we use the similarity difference between \texttt{done} and \texttt{ongoing} prompts:
\begin{equation}
    \begin{aligned}
        c_{t,k}=\sigma(\alpha(\langle f_t,e_k^{done}\rangle-\langle f_t,e_k^{ongoing}\rangle))\in[0,1]
    \end{aligned}
\end{equation}
where $\sigma(\cdot)$ is the sigmoid function and $\alpha$ controls the sharpness. We then define a scalar progress variable $p_t \in [0,1]$ based on the most probable phase and its completion:
\begin{equation}
    \begin{aligned}
        k_t^*=\arg\max_k \rho_{t,k},~~p_t=\frac{k_t^*+c_{t,k_t^*}}{K}.
    \end{aligned}
\end{equation}
Finally, the conditioning vector $z_t$ as shown in Eq \ref{eq:iv.a2} passed to the diffusion policy is $z_t=[h_t,\rho_t,p_t]\in\mathbb{R}^{512+K+1}$.
This construction is entirely zero-shot, with all phase and progress signals derived from natural language prompts and a frozen CLIP model without any manual phase labels.
By introducing the phase and progress head, diffusion policies no longer need to implicitly discover task phases purely from raw visual–proprioceptive embeddings; it can leverage the language-defined phases and progress signals as a weak supervisory scaffold, which empirically stabilizes training and improves downstream tasks' performance.

\section{Experiments and Analysis}

In this section, we evaluate the performance of the proposed framework across two tasks, each accompanied by two complementary experimental settings designed to assess different aspects of the system. We compare FALCON against the baseline methods and conduct ablation studies to examine the contribution of each component. Our analysis aims to provide a comprehensive understanding of the framework’s effectiveness.

\begin{figure*}[t]
    \centering
    \includegraphics[width=1\textwidth]{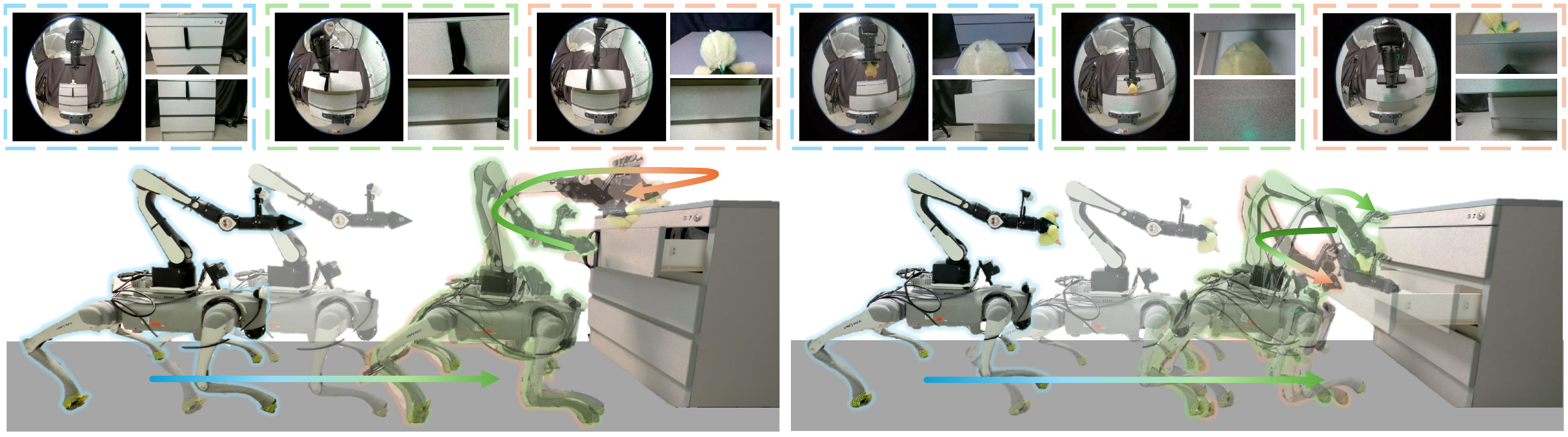}
    \caption{Overview of the manipulation tasks. The task on the left shows the quadruped–manipulator system approaching the drawer, adjusting its base pose for optimal arm reachability, opening the drawer, and placing a toy inside. The task on the right illustrates a complementary scenario in which the robot navigates to the drawer while holding a toy, places the toy into the already opened drawer, and then closes the drawer using pose-assisted whole-body manipulation. The top row shows synchronized egocentric camera observations for each stage, color-coded to match the corresponding robot pose, while the bottom row depicts the whole-body motions generated by the decoupled locomotion and manipulation policies.}
    \label{fig:overview}
\end{figure*}

\subsection{Experiments Design}
This section presents the experimental design used to systematically evaluate the proposed decoupled loco-manipulation framework. We consider two complementary experimental settings to assess both overall task performance and subsystem-level coordination under identical training conditions. 
The first experiment focuses on navigation and object pickup and placement, where the proposed method is quantitatively compared against representative baseline approaches. The second experiment investigates collaboration behavior in a human-in-the-loop setting, in which either the locomotion or manipulation policy is replaced by human teleoperation, allowing us to analyze how each subsystem performs when paired with a human-controlled counterpart. Together, these experiments evaluate the policies on two tasks, namely the \textit{precise manipulation task} and the \textit{mobile manipulation task}, and provide a comprehensive assessment of robustness, coordination capability, and practical deployment potential.

\subsubsection{Navigation and Pickup/Placement}
The first experiment evaluates the coordination effectiveness of the decoupled policies in a whole-body manipulation context. In this setting, the robot must navigate toward a designated target location and adjust its base pose to support the mounted arm in performing object pickup and placement. This experiment explicitly assesses the ability of the locomotion and manipulation subsystems to operate in a tightly coordinated manner, with the legged base providing appropriate positioning and stabilization to enable accurate and reliable manipulation.

The performance of the system is primarily measured through the overall task success rate and the consistency of coordinated behavior between the base and the manipulator throughout the navigation–manipulation sequence. These metrics allow us to quantify how effectively the two subsystems interact under the proposed decoupled control framework.

\subsubsection{Subsystem with Human-in-the-loop}

To demonstrate the advantages of the decoupled visuomotor policies, we conduct a human-in-the-loop experiment in which either the arm policy or the locomotion policy is disabled and replaced with real-time human teleoperation. This setup allows a human operator to intervene on only one subsystem without assuming control of the entire robot, enabling direct collaboration between a human-controlled subsystem and an autonomous one. 
The experiment considers two complementary modes: (i) human-controlled locomotion with autonomous manipulation, and (ii) autonomous locomotion with human-controlled manipulation. 
These modes allow us to evaluate how each subsystem performs when paired with human-generated behavior rather than its learned counterpart.

To avoid bias from operator familiarity, five volunteers participate in the study. Each volunteer performs two trials per mode, giving a total of ten trials for each mode (tele-base and tele-arm). 
The robot begins from varied initial positions across trials, ensuring that performance differences arise from subsystem interactions rather than specific starting states.

In the first mode, the human operator drives the legged base toward the target, after which the learned manipulation policy performs the pickup or placement behavior. 
This setting evaluates the manipulation subsystem’s robustness to non-ideal base approaches, such as inconsistent stopping positions or noisy trajectories introduced by human control. 
In the second mode, the locomotion policy autonomously positions the robot near the target, while the human operator controls the arm to complete the manipulation task. 
This mode assesses whether the locomotion subsystem provides stable, repeatable, and manipulation-friendly base configurations that support fine-grained human control.

Across both modes, we evaluate task success, base–arm coordination quality, and subsystem robustness when interacting with human-generated motion. 
By introducing natural human variability into the control loop, this experiment offers a meaningful assessment of the adaptability and practical usability of the proposed decoupled framework in real-world collaborative scenarios.

\subsection{Real Robot System Deployment}
On the real robot platform, we deploy our decoupled visuomotor policy and several baseline policies on a quadruped mobile manipulator composed of a Unitree Go2 base and an Airbot robotic arm. The system instantiates the tri-camera configuration described in Sec.~\ref{subsec:loco_manipulation_system_setup} using a wrist-mounted Intel RealSense D435i, a head-mounted Intel RealSense D455, and a body-mounted GoPro Hero~11 with a fisheye lens. All cameras operate in RGB-only mode, and no depth streams are used. The GoPro stream runs at 60\,Hz with a resolution of $640 \times 480$ pixels, while the RealSense D435i and D455 streams run at 15\,Hz with a resolution of $1280 \times 720$ pixels. All camera streams are bridged into ROS~2 image topics with precise timestamps to enable cross-view synchronization.

We adopt an offboard inference setup in which all policies run on a workstation equipped with an NVIDIA RTX~4090 GPU. The workstation communicates with the quadruped base over a wired network link to ensure low-latency command streaming, while the arm is controlled via a CAN interface using the official vendor SDK. Inference for the deployed visuomotor policy is performed at 10\,Hz, producing action chunks that are then executed by separate low-level controllers for the base and the arm. On the Go2 base, an RL low-level controller runs at 50\,Hz, while an inner joint-level control loop operates at 200\,Hz and issues new joint commands every 5\,ms; each 50\,Hz policy output is linearly interpolated and applied over four consecutive 200\,Hz control cycles. For the arm, the policy outputs desired end-effector positions in Cartesian space, and a low-level controller running at 100\,Hz interpolates these targets and converts them into joint commands for execution.

\begin{figure*}[t]
    \centering
    \includegraphics[width=\textwidth]{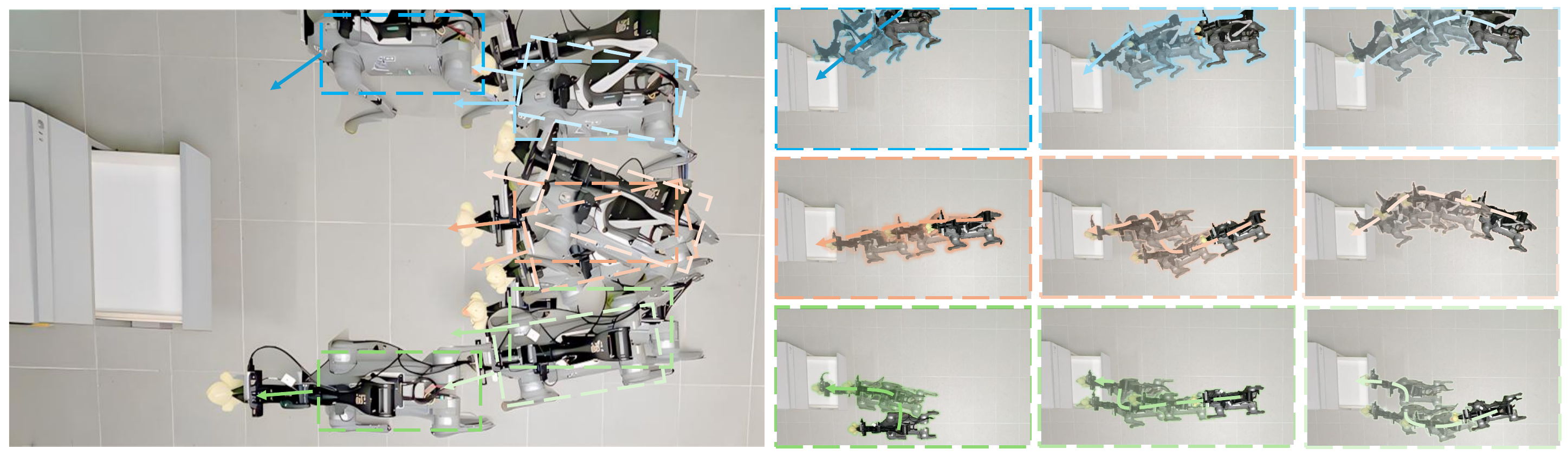}
    \caption{Initial robot positions and example navigation trajectories for Task~2. The robot begins from multiple starting locations across the workspace and navigates toward the opened drawer. The color-coding indicates the three evaluation regions: green for the left area, orange for the center area, and blue for the right area. Note that the training data are primarily collected from the center region, making the left and right regions predominantly out-of-distribution (OOD) during evaluation.}
    \label{fig:navigation}
\end{figure*}

\subsection{Baselines}

To evaluate the robustness of the proposed framework, we compare it against three baseline methods: a centralized diffusion policy (CDP)~\cite{chi2025diffusion}, action chunking with transformers (ACT)~\cite{zhao2023learning,fu2024mobile}, and LatentToM~\cite{he2025latent}. For our method, the visuomotor policy is trained using phase instructions provided through textual prompts, together with a coordination-aware contrastive loss that sharpens the latent representation. All approaches are trained on the same set of 50 teleoperated demonstrations collected on the real robot, and experiments are conducted under identical training conditions to ensure a fair comparison. To improve generalization, the robot’s initial position and heading are randomized within a predefined region for each task.

For the Subsystem with Human-in-the-loop evaluation, the centralized CDP policy cannot be decomposed into separate locomotion and manipulation components and is therefore incompatible with this setting. Consequently, for this experiment we compare only against LatentToM, which, like our method, employs a decoupled control structure.

\subsection{Precise Manipulation Task}

\begin{figure}[h]
    \centering
    \includegraphics[width=0.9\columnwidth]{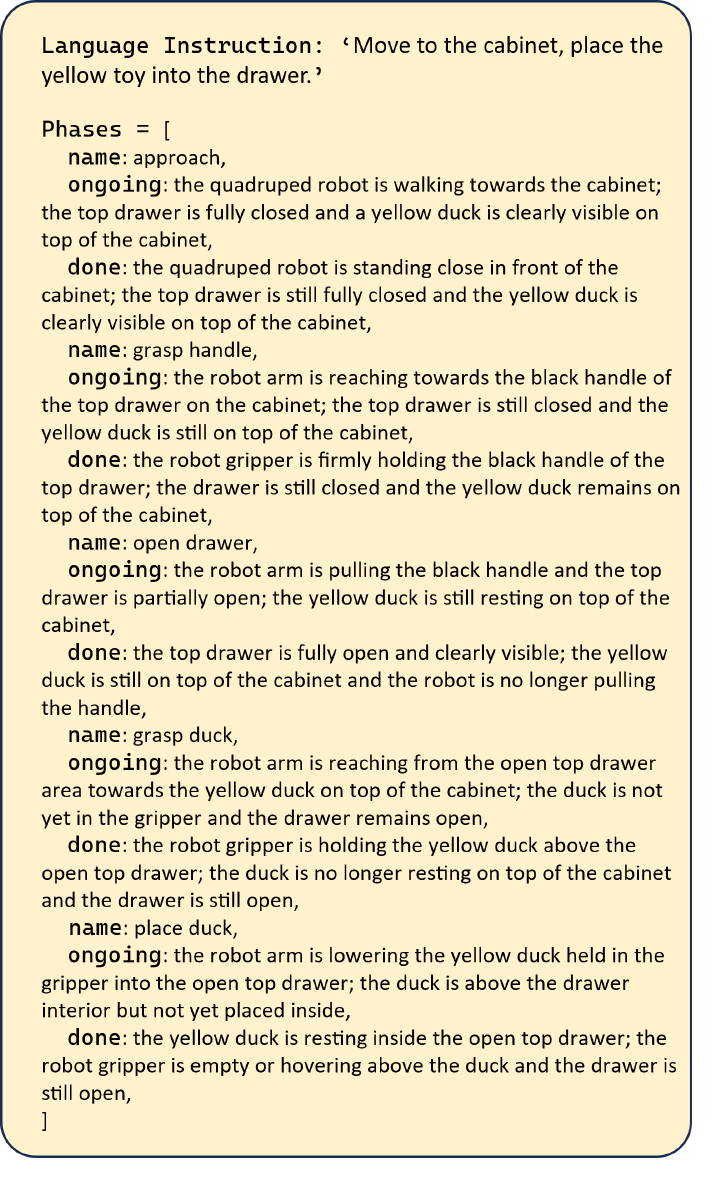}
    \caption{Precise manipulation task language instruction and phase prompts.}
    \label{fig:task1_prompt}
\end{figure}

In this task (Task 1), the robot is required to navigate toward a cabinet with three closed drawers, open the top of them, pick up a toy placed on top of the cabinet, and finally place the toy inside the top drawer (shown in Fig.\ref{fig:overview} left). 
To accomplish this, the robot must first move to the required position and then operate the manipulator to interact with the drawer handle and execute the opening action before proceeding with object pickup and placement. 
This setting emphasizes fine manipulation and perception-guided control in a constrained workspace.
During data collection, we introduce a degree of randomness to robot’s initial position setting in the hope of improving generalization.
The drawer is initialized in a fully closed configuration, requiring the policy to infer and execute precise pulling motions for successful opening. 
In addition, the toy position on the top of the cabinet is varied to further increase task diversity.
The phase-specific language prompt and the overall task's language instruction used during Task 1 training are shown in Fig ~\ref{fig:task1_prompt}.

\subsubsection{Policies Evaluation}

During evaluation, the robot’s initial pose is randomly sampled within the room (as shown in Fig.~\ref{fig:navigation}). 
We assess task performance in a stage-wise manner by decomposing the procedure into four sequential stages: (1) navigating to a manipulation-capable position in front of the closed drawer, (2) opening the drawer through coordinated interaction with the handle, (3) picking up the toy, and (4) placing the toy inside the drawer. This structured breakdown allows for independent evaluation of navigation precision, drawer-opening reliability, grasping success, and placement accuracy. Each method is evaluated over 15 trials with varied initial positions (5 trials each for the center, left, and right regions), providing a comprehensive measure of robustness and reproducibility.

\begin{figure*}[t]
    \centering
    \includegraphics[width=\textwidth]{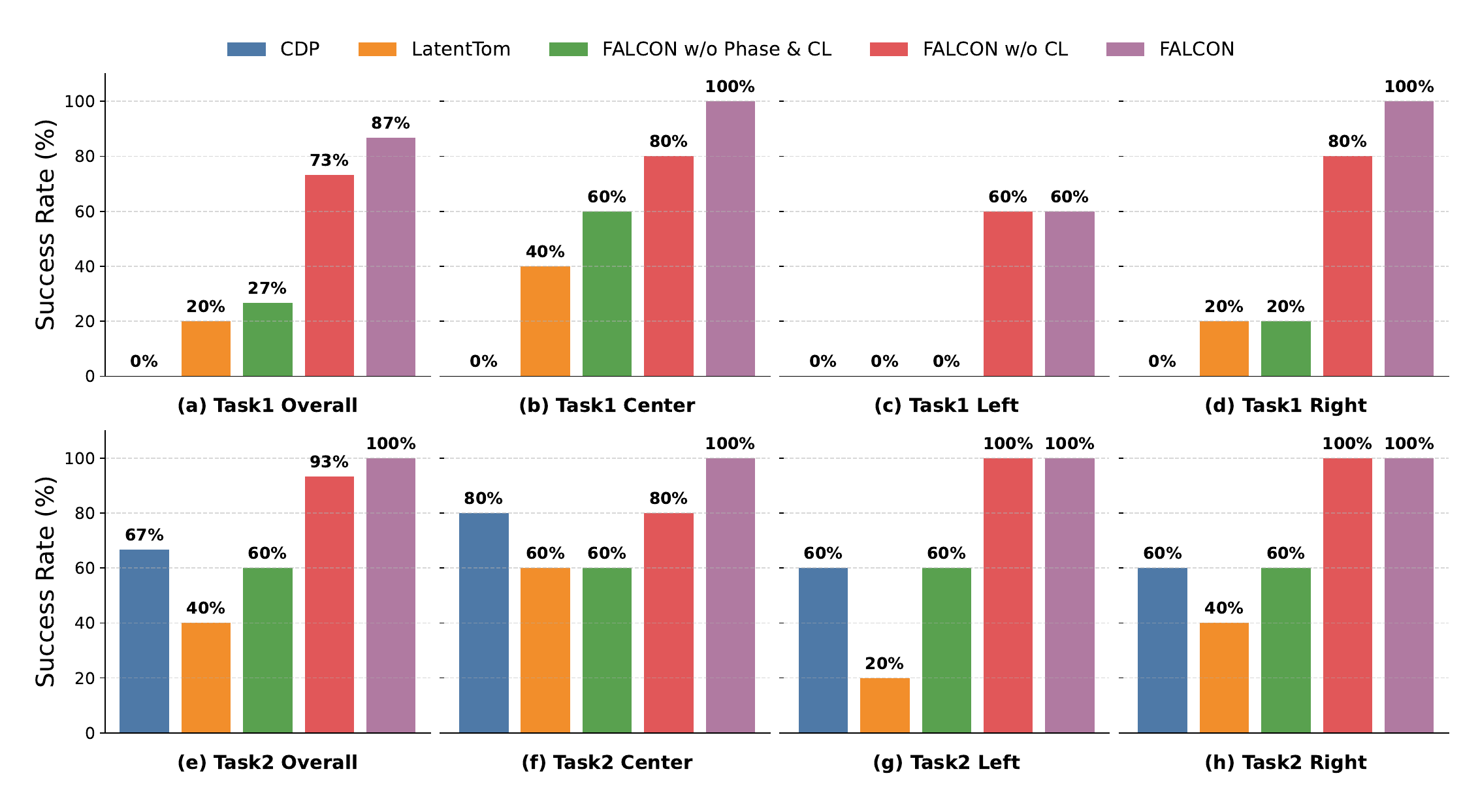}
    \caption{Success rates for Task 1 (precise manipulation) and Task 2 (navigation and placement) across different initial positions. Each subplot reports the performance of the baselines (CDP and LatentTOM), ablated variants of our method (FALCON w/o Phase \& CL and FALCON w/o CL), and the full FALCON framework. The results show that FALCON consistently achieves the highest success rates across all task settings and initial configurations. The ACT baseline is omitted from this figure because it achieves a zero percent task-level success rate in all experiments; however, its stage-wise performance can be found in Table~\ref{tab:phase_success}.}
    \label{fig:task12_success_rates}
\end{figure*}

As shown in Fig.~\ref{fig:task12_success_rates} and Table~\ref{tab:phase_success}, FALCON outperformed than the baselines.
One very interesting and unusual phenomenon we observed is that decentralized methods (LatentToM and FALCON) significantly outperformed centralized methods (CDP and ACT), where ACT failed to successfully complete the entire task process even once.
This phenomenon arises because the task requires both successful navigation (including reaching a feasible operating pose) and precise manipulation (such as grasping drawer handles and picking up toys). 
For locomotion, our quadruped base is controlled via high-level velocity commands, which makes precise repositioning substantially more challenging than for a wheeled base with more straightforward kinematics. 
Meanwhile, the manipulator has a very limited workspace, so the accuracy of the quadruped’s motion has a disproportionately large impact on the overall success rate. 
This strong asymmetry between the quadruped and the manipulator makes it difficult for a centralized policy to learn both skills effectively at once. 
In contrast, our decoupled framework separates the control objectives of the base and the manipulator, allowing each subsystem to specialize in its own function while still achieving consistent end-to-end task performance.

By comparing the full phase–progress conditioning against a baseline that uses only the global embedding as input to the diffusion policy, we observe a substantial performance drop when phase and progress information are removed. 
This indicates that the language-defined phase and progress signals provide a useful inductive bias for optimization. 
By exposing a coarse phase distribution and a continuous notion of task completion, they decompose an otherwise highly multi-modal action distribution into simpler conditional distributions, which significantly stabilizes and accelerates diffusion policy training.
In our experiments, the progress variable exhibits clear and relatively accurate variation over the course of each trial. 
We believe this supplies a coarse, time-like signal to the policy that is sufficient for the diffusion model to learn a reliable mapping from observations to actions.
Finally, we found that the contrastive loss yielded a larger performance gain in Task 1 than in Task 2. 
We believe this is because Task 1 remains more easier to stay in in-distribution (InD) during execution, whereas Task 2 encounters more out-of-distribution conditions. 
By design, contrastive learning is particularly effective when the test-time distribution closely matches the training data, which explains its stronger impact in the more InD setting of Task 1.

\subsubsection{Subsystem with Human-in-the-loop Evaluation}

The results of the human-in-the-loop experiments are shown in Fig.~\ref{fig:teleop_success}. Our FALCON framework achieves a 100\% success rate across all ten trials. To separately evaluate the contributions of the two subsystems, we decompose task accomplishment into two criteria: (1) \emph{arm task success}, defined as the ability to open the drawer, pick up the toy, and place it inside the drawer, and (2) \emph{legged-base task success}, defined as the ability to navigate to a valid pose centered in front of the drawer.

Detailed quantitative results are provided in Table~\ref{tab:teleop_success}. As expected, the teleoperated subsystem always achieves 100\% success, confirming that any failures originate from the policy-controlled subsystem. When the base is teleoperated, FALCON achieves higher manipulation success than LatentToM, indicating that its manipulation policy remains reliable even under human-induced variations in base positioning. This demonstrates that the shared latent representation learned through phase–progress modeling and coordination-aware contrastive learning provides each subsystem with sufficient contextual awareness to operate robustly when paired with a human-controlled counterpart. These results highlight the robustness, modularity, and practical utility of the proposed decoupled framework in mixed-autonomy scenarios.

\subsection{Mobile Manipulation Task}

\begin{figure}[h]
    \centering
    \includegraphics[width=0.9\columnwidth]{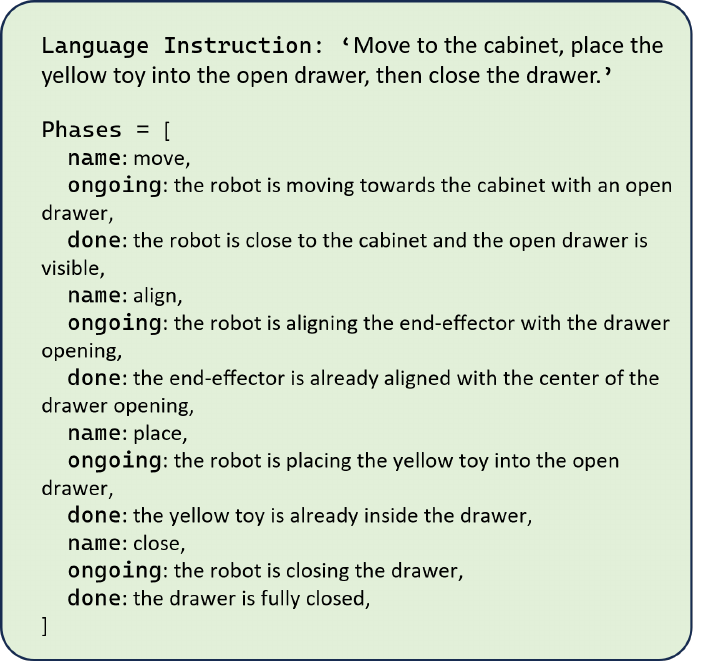}
    \caption{Mobile manipulation task language instruction and phase prompts.}
    \label{fig:task2_prompt}
\end{figure}

In the second task, we focus on evaluating navigation performance and the coordination between the mobile base and the manipulator. The robot starts with a toy held in its gripper and is required to navigate to an opened drawer, place the toy inside, and subsequently close the drawer. 
During data collect stage,  both the robot’s initial position and heading are randomized within a predefined region.
In addition, the drawer is initialized at varying opening degrees (e.g., half-open and fully open) to introduce diverse manipulation scenarios.
The phase-specific language prompt and the overall task's language instruction used during task 1 training are shown in Fig ~\ref{fig:task2_prompt}.

During evaluation, unlike the first task, we assess the policy under an out-of-distribution (OOD) setting in which the robot’s initial position is sampled from a larger area, as illustrated in Fig.~\ref{fig:navigation}. 
For quantitative analysis, the experimental workspace is partitioned into three regions (left, center, and right), and the overall task is decomposed into three sequential stages and use them as the metrics: (1) navigating to a manipulation-capable position in front of the drawer, (2) placing the toy inside the drawer, and (3) closing the drawer. This stage-based decomposition enables separate assessment of navigation performance, manipulation accuracy, and whole-body coordination.

\begin{table*}[t]
\centering
\caption{Stage-wise success rates (\%) for Task 1 (precise drawer manipulation) and Task 2 (navigation and placement) across Center, Left, and Right initial configurations. For each method, performance is reported separately for all task stages, including navigation, intermediate drawer interactions, and object pickup or placement. The overall task completion success rates are provided in Fig.~\ref{fig:task12_success_rates}. This table highlights differences in subsystem coordination ability between the baseline methods, ablated variants of FALCON, and the full FALCON framework. Consistent improvements are observed in both early navigation stages and later manipulation stages, demonstrating the effectiveness of the phase–progress modeling and the coordination-aware contrastive learning components.}
\label{tab:phase_success}
\setlength{\tabcolsep}{7pt}
\small
\begin{tabular}{lllcccccc}
\toprule
\multirow{2}{*}{Task} & \multirow{2}{*}{Region} & \multirow{2}{*}{Stage} 
& CDP & ACT & LatentToM 
& FALCON w/o Phase \& CL & FALCON w/o CL & FALCON (ours) \\
\cmidrule(lr){4-9}
 &  &  & \multicolumn{6}{c}{Success Rate (\%)} \\
\midrule

\multirow{12}{*}{Task 1}
 & \multirow{4}{*}{Center}
   & Navigation        & 40 & 40 & 100 & 100 & 100 & \textbf{100} \\
 & & Open Drawer       & 0  & 0  & 60  & 80  & 100 & \textbf{100} \\
 & & Pick up Toy       & 0  & 0  & 40  & 60  & 80  & \textbf{100} \\
 & & Place Toy         & 0  & 0  & 40  & 60  & 80  & \textbf{100} \\
\cmidrule(lr){2-9}

 & \multirow{4}{*}{Left}
   & Navigation        & 0  & 20 & 80  & 100 & 100 & \textbf{100} \\
 & & Open Drawer       & 0  & 0  & 20  & 20  & 60  & \textbf{80} \\
 & & Pick up Toy       & 0  & 0  & 0   & 0   & 60  & \textbf{60} \\
 & & Place Toy         & 0  & 0  & 0   & 0   & 60  & \textbf{60} \\
\cmidrule(lr){2-9}

 & \multirow{4}{*}{Right}
   & Navigation        & 40 & 20 & 80  & 100 & 100 & \textbf{100} \\
 & & Open Drawer       & 0  & 0  & 40  & 20  & 100 & \textbf{100} \\
 & & Pick up Toy       & 0  & 0  & 20  & 20  & 80  & \textbf{100} \\
 & & Place Toy         & 0  & 0  & 20  & 20  & 80  & \textbf{100} \\
\midrule

\multirow{9}{*}{Task 2}
 & \multirow{3}{*}{Center}
   & Navigation        & 80 & 100 & 100 & 60  & 80  & 80 \\
 & & Place Toy         & 80 & 80  & 60  & 60  & \textbf{100} & \textbf{100} \\
 & & Close Drawer      & 80 & 0   & 60  & 60  & 80  & \textbf{100} \\
\cmidrule(lr){2-9}

 & \multirow{3}{*}{Left}
   & Navigation        & 20 & 60 & 20  & 60  & \textbf{100} & 60 \\
 & & Place Toy         & 60 & 40 & 20  & 60  & \textbf{100} & \textbf{100} \\
 & & Close Drawer      & 60 & 0  & 20  & 60  & \textbf{100} & \textbf{100} \\
\cmidrule(lr){2-9}

 & \multirow{3}{*}{Right}
   & Navigation        & 20 & 60 & 60  & 20  & \textbf{100} & 80 \\
 & & Place Toy         & 60 & 60 & 60  & 60  & \textbf{100} & \textbf{100} \\
 & & Close Drawer      & 60 & 0  & 40  & 60  & \textbf{100} & \textbf{100} \\
\bottomrule
\end{tabular}
\end{table*}

\subsubsection{Policies Evaluation}

In the first experiment of this task, the proposed framework and all baseline methods are evaluated over 15 trials with different initial robot's positions and headings, while the drawer opening degree is also varied across trials. This setup provides a comprehensive assessment of robustness and generalization. Task success is defined as the toy being successfully placed inside the drawer followed by the drawer being fully closed. The training dataset is primarily collected with the robot starting from the center region. Consequently, many initial poses in the left and right regions do not appear during training and can therefore be regarded as out-of-distribution (OOD) scenarios.

As shown in Fig.~\ref{fig:task12_success_rates}, the overall success rates demonstrate that FALCON consistently outperforms all baseline methods, achieving 100\% success across all trials. Detailed stage-wise results are reported in Table~\ref{tab:phase_success}. FALCON also achieves the strongest performance in the left and right starting regions, which are rarely encountered during training. Notably, the version of FALCON without the contrastive loss attains performance comparable to the full model and even exceeds it in the navigation stage, as indicated in Table~\ref{tab:phase_success}. Although the contrastive loss is intended to encourage better alignment between the decoupled base and arm policies, it also introduces stronger coupling between them. This tighter coupling can reduce the generalization capability of the mobile base in OOD conditions because the base becomes more dependent on manipulation-specific latent cues. Removing the contrastive loss allows the locomotion subsystem to retain greater autonomy and adaptability when operating from unfamiliar initial states. However, the full FALCON model still achieves a slightly higher overall success rate. Even in cases where the base does not navigate to an ideal manipulation position, the arm policy is able to complete the remaining stages of the task, which compensates for minor navigation errors. This observation highlights the robustness of the proposed decoupled design and shows that, despite occasional navigation deviations, the coordinated latent representation enables reliable whole-body task completion.

ACT achieves relatively high success rates in the navigation and toy-placement stages; however, it consistently fails in the drawer-closing stage, which requires coordinated whole-body motion between the base and the arm. We observe that when the base tilts forward, the ACT-controlled arm does not retract the end-effector appropriately, resulting in collisions with the opened drawer edge. This behavior leads to a zero success rate for ACT in the drawer-closing stage.
LatentToM performs well in the navigation stage under in-distribution (ID) conditions, owing to its weaker implicit coupling between the base and arm policies. Its performance, however, drops substantially during the manipulation stages. The centralized diffusion policy shows comparable performance in ID scenarios but suffers a noticeable decline under OOD conditions.
Finally, the ablation study shows that the inclusion of the language-defined phase and progress head consistently improves performance across all stages, highlighting its importance for enabling robust, generalizable task execution.

\begin{table*}[t]
\centering
\caption{Subsystem success rates under human teleoperation for Task~1 and Task~2. For each task, we evaluate two modes: Tele-base, where a human operator drives the quadruped base while the manipulation policy completes the task, and Tele-arm, where autonomous locomotion positions the robot and the human teleoperates the arm. The reported success rates show how well each subsystem performs when paired with a human-controlled counterpart, highlighting the coordination robustness enabled by our decoupled visuomotor framework.}
\label{tab:teleop_success}
\setlength{\tabcolsep}{8pt}
\small
\begin{tabular}{l l cc|cc}
\toprule
Task & Mode 
& \multicolumn{2}{c|}{LatentToM} 
& \multicolumn{2}{c}{FALCON (ours)} \\
 &  & Base Success & Arm Success & Base Success & Arm Success \\
\midrule

\multirow{2}{*}{Task~1}
 & Tele-base & 100.0\% & 60.0\%  & 100.0\% & \textbf{80.0\%} \\
 & Tele-arm  & 90.0\%  & 100.0\% & \textbf{100.0\%}           & 100.0\% \\
\midrule

\multirow{2}{*}{Task~2}
 & Tele-base & 100.0\% & 80.0\%  & 100.0\% & \textbf{100.0\%} \\
 & Tele-arm  & 70.0\%  & 100.0\% & \textbf{90.0\%}           & 100.0\% \\
\bottomrule
\end{tabular}
\end{table*}

\begin{figure}[h]
    \centering
    \includegraphics[width=1\columnwidth]{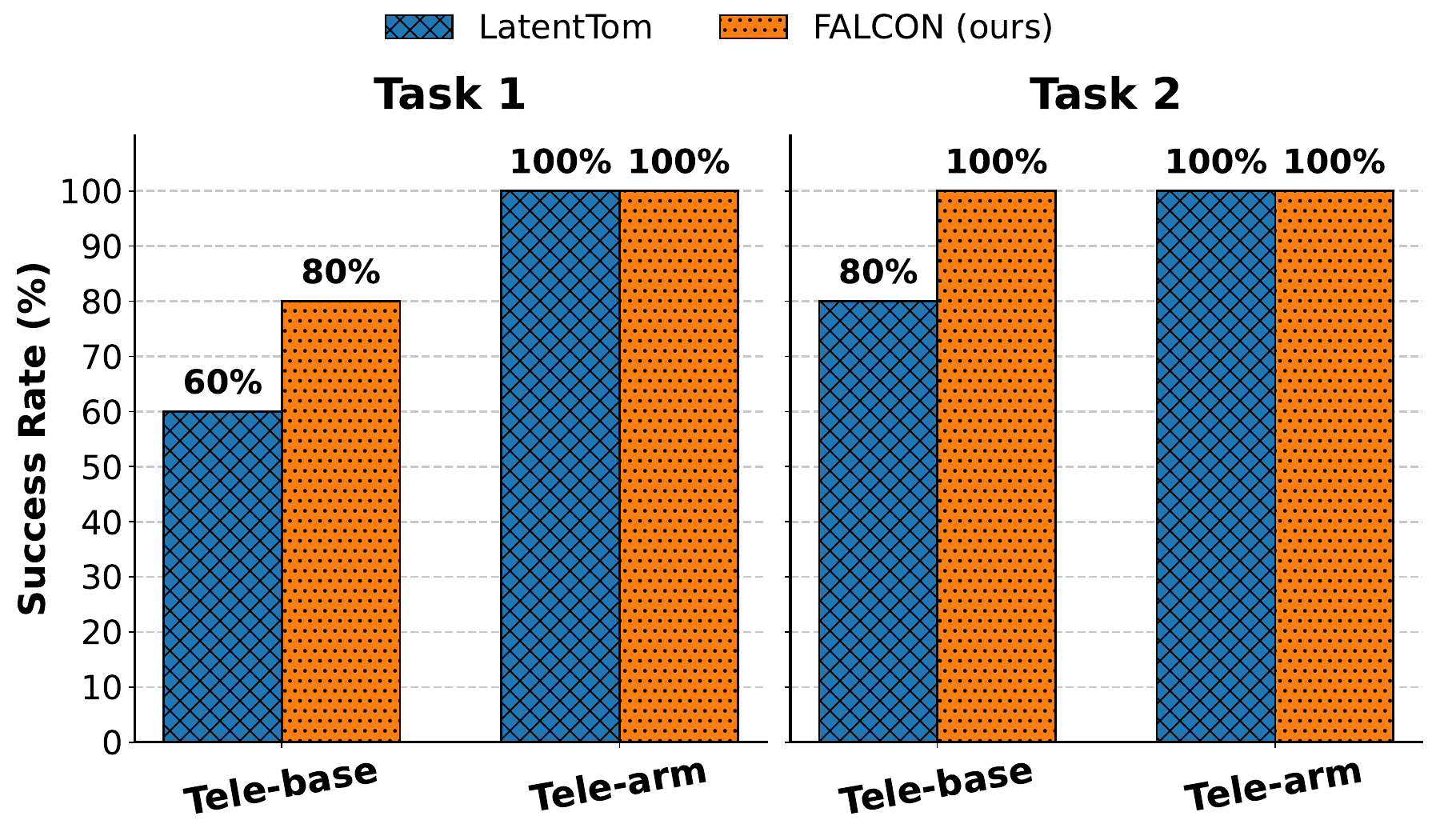}
    \caption{Teleoperation subsystem success rate comparison for LatentToM and FALCON (Task1 \& Task 2).}
    \label{fig:teleop_success}
\end{figure}

\subsubsection{Subsystem with Human-in-the-loop Evaluation}

As shown in Fig.~\ref{fig:teleop_success}, our FALCON framework outperforms LatentToM in overall success rate.
We further decompose task accomplishment into two subsystem-specific criteria: (1) \emph{arm task success}, defined as the toy being placed inside the drawer and the drawer being fully closed, and (2) \emph{legged base task success}, defined as the base navigating to a valid position centered in front of the drawer. The detailed results in Table~\ref{tab:teleop_success} show that the teleoperated subsystem always achieves a 100\% success rate, as expected. More importantly, the subsystem controlled by FALCON (when coordinating with the human-operated counterpart) exhibits substantially better robustness than LatentToM.
The only failure cases observed for FALCON in the tele-arm condition occur when the teleoperator unintentionally moves the end effector into a position that obstructs the fisheye camera, which disrupts the legged-base policy and causes the body to collide with the opened drawer. However, these collisions do not prevent the toy from being placed inside the drawer or the drawer from being closed under teleoperated arm control, and the overall task is still successfully completed in those trials.

\section{Discussion on Limitations and Future Work}

This section summarizes the main limitations of the current system and outlines directions for future research.

First, the coordination mechanism relies on manually crafted textual descriptions for each task phase, which are interpreted by a frozen CLIP model to infer phase and progress. Although this enables phase-aware reasoning without explicit annotations, it introduces sensitivity to prompt quality and to domain differences between internet-scale training data and robot environments. Subtle visual variations or cluttered scenes may therefore reduce estimation reliability. Future work will explore the use of large language models to automatically generate or refine phase cues across larger and more diverse task sets, with the aim of improving accuracy while retaining the zero-shot generalization benefits of the current approach.

Second, the tasks used in our experiments cover navigation-driven and precise, body-assisted manipulation scenarios, but they do not capture the full complexity or long-horizon structure of real-world activities. Extending the evaluation to more diverse, multi-step tasks with richer environmental variation and higher semantic complexity will provide a deeper understanding of how the proposed framework scales to practical loco-manipulation settings.

Finally, an important direction is to study how FALCON generalizes across different robot embodiments and object categories. The decoupled policy structure suggests that locomotion and manipulation modules can be transferred or adapted with minimal additional data, and potentially even without retraining. Investigating this capability across broader loco-manipulation families and open-vocabulary goals will be critical for advancing toward more general-purpose robotic systems.

\section{Conclusion}

In this work, we introduced FALCON, a decoupled modular diffusion-policy architecture for loco-manipulation that leverages a foundation model to coordinate subsystem behavior. 
By explicitly separating locomotion and manipulation into a velocity diffusion policy atop a low-level RL controller for the quadruped and a position diffusion policy for the mounted manipulator through IK, our system allows each subsystem to operate in its natural observation and control space, mitigating the instability and observation-mismatch failure modes common in centralized whole-body policies. 
A frozen CLIP backbone encodes global RGB observations, robot states, and language cues into a shared latent embedding that conditions both policies, enabling coordinated behavior despite decentralized control. 
Through a set of challenging loco-manipulation tasks requiring long-horizon navigation, precise end-effector placement, and tight base–arm coordination, we demonstrate the effectiveness and robustness of FALCON. 
Our results show that FALCON surpasses the centralized and decentralized baselines. 
In particular, for base locomotion, our method achieves accurate repositioning with limited data, providing a reliable foundation for the successful operation of the mounted manipulator.

\bibliographystyle{IEEEtran}
\bibliography{ref}

@article{long2023hybrid,
  title={Hybrid internal model: Learning agile legged locomotion with simulated robot response},
  author={Long, Junfeng and Wang, Zirui and Li, Quanyi and Gao, Jiawei and Cao, Liu and Pang, Jiangmiao},
  journal={arXiv preprint arXiv:2312.11460},
  year={2023}
}

@article{he2025latent,
  title={Latent Theory of Mind: A Decentralized Diffusion Architecture for Cooperative Manipulation},
  author={He, Chengyang and Camps, Gadiel Sznaier and Liu, Xu and Schwager, Mac and Sartoretti, Guillaume},
  journal={arXiv preprint arXiv:2505.09144},
  year={2025}
}

@article{schulman2017proximal,
  title={Proximal policy optimization algorithms},
  author={Schulman, John and Wolski, Filip and Dhariwal, Prafulla and Radford, Alec and Klimov, Oleg},
  journal={arXiv preprint arXiv:1707.06347},
  year={2017}
}

@article{chi2025diffusion,
  title={Diffusion policy: Visuomotor policy learning via action diffusion},
  author={Chi, Cheng and Xu, Zhenjia and Feng, Siyuan and Cousineau, Eric and Du, Yilun and Burchfiel, Benjamin and Tedrake, Russ and Song, Shuran},
  journal={The International Journal of Robotics Research},
  volume={44},
  number={10-11},
  pages={1684--1704},
  year={2025},
  publisher={Sage Publications Sage UK: London, England}
}

@article{firoozi2025foundation,
  title={Foundation models in robotics: Applications, challenges, and the future},
  author={Firoozi, Roya and Tucker, Johnathan and Tian, Stephen and Majumdar, Anirudha and Sun, Jiankai and Liu, Weiyu and Zhu, Yuke and Song, Shuran and Kapoor, Ashish and Hausman, Karol and others},
  journal={The International Journal of Robotics Research},
  volume={44},
  number={5},
  pages={701--739},
  year={2025},
  publisher={SAGE Publications Sage UK: London, England}
}

@article{black2024pi0,
  title   = {${\pi}_{0}$: A Vision--Language--Action Flow Model for General Robot Control},
  author  = {Black, Kevin and Brown, Noah and Driess, Danny and Esmail, Adnan and Equi, Michael and Finn, Chelsea and Fusai, Niccolo and Groom, Lachy and Hausman, Karol and Ichter, Brian and others},
  journal = {arXiv preprint arXiv:2410.24164},
  year    = {2024},
  doi     = {10.48550/arXiv.2410.24164},
  note    = {CoRR, abs/2410.24164}
}

@misc{ho2021sim2real,
  title        = {Toward Generalized Sim-to-Real Transfer for Robot Learning},
  author       = {Ho, Daniel and Rao, Kanishka},
  howpublished = {\url{https://research.google/blog/toward-generalized-sim-to-real-transfer-for-robot-learning/}},
  note         = {Google Research Blog},
  year         = {2021},
  month        = jun
}

@article{khan2025foundation,
  title={Foundation model driven robotics: A comprehensive review},
  author={Khan, Muhammad Tayyab and Waheed, Ammar},
  journal={arXiv preprint arXiv:2507.10087},
  year={2025}
}

@article{meng2025preserving,
  title={Preserving and combining knowledge in robotic lifelong reinforcement learning},
  author={Meng, Yuan and Bing, Zhenshan and Yao, Xiangtong and Chen, Kejia and Huang, Kai and Gao, Yang and Sun, Fuchun and Knoll, Alois},
  journal={Nature Machine Intelligence},
  pages={1--14},
  year={2025},
  publisher={Nature Publishing Group UK London}
}

@article{zeng2023large,
  title={Large language models for robotics: A survey},
  author={Zeng, Fanlong and Gan, Wensheng and Wang, Yongheng and Liu, Ning and Yu, Philip S},
  journal={arXiv preprint arXiv:2311.07226},
  year={2023}
}

@inproceedings{radford2021learning,
  title={Learning transferable visual models from natural language supervision},
  author={Radford, Alec and Kim, Jong Wook and Hallacy, Chris and Ramesh, Aditya and Goh, Gabriel and Agarwal, Sandhini and Sastry, Girish and Askell, Amanda and Mishkin, Pamela and Clark, Jack and others},
  booktitle={International conference on machine learning},
  pages={8748--8763},
  year={2021},
  organization={PmLR}
}

@inproceedings{kirillov2023segment,
  title={Segment anything},
  author={Kirillov, Alexander and Mintun, Eric and Ravi, Nikhila and Mao, Hanzi and Rolland, Chloe and Gustafson, Laura and Xiao, Tete and Whitehead, Spencer and Berg, Alexander C and Lo, Wan-Yen and others},
  booktitle={Proceedings of the IEEE/CVF international conference on computer vision},
  pages={4015--4026},
  year={2023}
}

@article{nair2022r3m,
  title={R3m: A universal visual representation for robot manipulation},
  author={Nair, Suraj and Rajeswaran, Aravind and Kumar, Vikash and Finn, Chelsea and Gupta, Abhinav},
  journal={arXiv preprint arXiv:2203.12601},
  year={2022}
}

@inproceedings{grauman2022ego4d,
  title={Ego4d: Around the world in 3,000 hours of egocentric video},
  author={Grauman, Kristen and Westbury, Andrew and Byrne, Eugene and Chavis, Zachary and Furnari, Antonino and Girdhar, Rohit and Hamburger, Jackson and Jiang, Hao and Liu, Miao and Liu, Xingyu and others},
  booktitle={Proceedings of the IEEE/CVF conference on computer vision and pattern recognition},
  pages={18995--19012},
  year={2022}
}

@inproceedings{zhou2022detecting,
  title={Detecting twenty-thousand classes using image-level supervision},
  author={Zhou, Xingyi and Girdhar, Rohit and Joulin, Armand and Kr{\"a}henb{\"u}hl, Philipp and Misra, Ishan},
  booktitle={European conference on computer vision},
  pages={350--368},
  year={2022},
  organization={Springer}
}

@inproceedings{minderer2022simple,
  title={Simple open-vocabulary object detection},
  author={Minderer, Matthias and Gritsenko, Alexey and Stone, Austin and Neumann, Maxim and Weissenborn, Dirk and Dosovitskiy, Alexey and Mahendran, Aravindh and Arnab, Anurag and Dehghani, Mostafa and Shen, Zhuoran and others},
  booktitle={European conference on computer vision},
  pages={728--755},
  year={2022},
  organization={Springer}
}

@article{gu2021open,
  title={Open-vocabulary object detection via vision and language knowledge distillation},
  author={Gu, Xiuye and Lin, Tsung-Yi and Kuo, Weicheng and Cui, Yin},
  journal={arXiv preprint arXiv:2104.13921},
  year={2021}
}

@inproceedings{shah2023lm,
  title={Lm-nav: Robotic navigation with large pre-trained models of language, vision, and action},
  author={Shah, Dhruv and Osi{\'n}ski, B{\l}a{\.z}ej and Levine, Sergey and others},
  booktitle={Conference on robot learning},
  pages={492--504},
  year={2023},
  organization={PMLR}
}

@article{shafiullah2022clip,
  title={Clip-fields: Weakly supervised semantic fields for robotic memory},
  author={Shafiullah, Nur Muhammad Mahi and Paxton, Chris and Pinto, Lerrel and Chintala, Soumith and Szlam, Arthur},
  journal={arXiv preprint arXiv:2210.05663},
  year={2022}
}

@article{ahn2022can,
  title={Do as i can, not as i say: Grounding language in robotic affordances},
  author={Ahn, Michael and Brohan, Anthony and Brown, Noah and Chebotar, Yevgen and Cortes, Omar and David, Byron and Finn, Chelsea and Fu, Chuyuan and Gopalakrishnan, Keerthana and Hausman, Karol and others},
  journal={arXiv preprint arXiv:2204.01691},
  year={2022}
}

@article{chowdhery2023palm,
  title={Palm: Scaling language modeling with pathways},
  author={Chowdhery, Aakanksha and Narang, Sharan and Devlin, Jacob and Bosma, Maarten and Mishra, Gaurav and Roberts, Adam and Barham, Paul and Chung, Hyung Won and Sutton, Charles and Gehrmann, Sebastian and others},
  journal={Journal of Machine Learning Research},
  volume={24},
  number={240},
  pages={1--113},
  year={2023}
}

@article{driess2023palm,
  title={Palm-e: An embodied multimodal language model},
  author={Driess, Danny and Xia, Fei and Sajjadi, Mehdi SM and Lynch, Corey and Chowdhery, Aakanksha and Wahid, Ayzaan and Tompson, Jonathan and Vuong, Quan and Yu, Tianhe and Huang, Wenlong and others},
  year={2023}
}

@article{wei2022chain,
  title={Chain-of-thought prompting elicits reasoning in large language models},
  author={Wei, Jason and Wang, Xuezhi and Schuurmans, Dale and Bosma, Maarten and Xia, Fei and Chi, Ed and Le, Quoc V and Zhou, Denny and others},
  journal={Advances in neural information processing systems},
  volume={35},
  pages={24824--24837},
  year={2022}
}

@article{brohan2022rt,
  title={Rt-1: Robotics transformer for real-world control at scale},
  author={Brohan, Anthony and Brown, Noah and Carbajal, Justice and Chebotar, Yevgen and Dabis, Joseph and Finn, Chelsea and Gopalakrishnan, Keerthana and Hausman, Karol and Herzog, Alex and Hsu, Jasmine and others},
  journal={arXiv preprint arXiv:2212.06817},
  year={2022}
}

@inproceedings{shridhar2022cliport,
  title={Cliport: What and where pathways for robotic manipulation},
  author={Shridhar, Mohit and Manuelli, Lucas and Fox, Dieter},
  booktitle={Conference on robot learning},
  pages={894--906},
  year={2022},
  organization={PMLR}
}

@inproceedings{shridhar2023perceiver,
  title={Perceiver-actor: A multi-task transformer for robotic manipulation},
  author={Shridhar, Mohit and Manuelli, Lucas and Fox, Dieter},
  booktitle={Conference on Robot Learning},
  pages={785--799},
  year={2023},
  organization={PMLR}
}

@inproceedings{zitkovich2023rt,
  title={Rt-2: Vision-language-action models transfer web knowledge to robotic control},
  author={Zitkovich, Brianna and Yu, Tianhe and Xu, Sichun and Xu, Peng and Xiao, Ted and Xia, Fei and Wu, Jialin and Wohlhart, Paul and Welker, Stefan and Wahid, Ayzaan and others},
  booktitle={Conference on Robot Learning},
  pages={2165--2183},
  year={2023},
  organization={PMLR}
}

@article{kim2024openvla,
  title={Openvla: An open-source vision-language-action model, 2024},
  author={Kim, Moo Jin and Pertsch, Karl and Karamcheti, Siddharth and Xiao, Ted and Balakrishna, Ashwin and Nair, Suraj and Rafailov, Rafael and Foster, Ethan and Lam, Grace and Sanketi, Pannag and others},
  journal={URL https://arxiv. org/abs/2406.09246},
  year={2024}
}

@article{reed2022generalist,
  title={A generalist agent},
  author={Reed, Scott and Zolna, Konrad and Parisotto, Emilio and Colmenarejo, Sergio Gomez and Novikov, Alexander and Barth-Maron, Gabriel and Gimenez, Mai and Sulsky, Yury and Kay, Jackie and Springenberg, Jost Tobias and others},
  journal={arXiv preprint arXiv:2205.06175},
  year={2022}
}

@inproceedings{o2024open,
  title={Open x-embodiment: Robotic learning datasets and rt-x models: Open x-embodiment collaboration 0},
  author={O’Neill, Abby and Rehman, Abdul and Maddukuri, Abhiram and Gupta, Abhishek and Padalkar, Abhishek and Lee, Abraham and Pooley, Acorn and Gupta, Agrim and Mandlekar, Ajay and Jain, Ajinkya and others},
  booktitle={2024 IEEE International Conference on Robotics and Automation (ICRA)},
  pages={6892--6903},
  year={2024},
  organization={IEEE}
}

@article{fu2024mobile,
  title={Mobile aloha: Learning bimanual mobile manipulation with low-cost whole-body teleoperation},
  author={Fu, Zipeng and Zhao, Tony Z and Finn, Chelsea},
  journal={arXiv preprint arXiv:2401.02117},
  year={2024}
}

@article{bjorck2025gr00t,
  title={Gr00t n1: An open foundation model for generalist humanoid robots},
  author={Bjorck, Johan and Casta{\~n}eda, Fernando and Cherniadev, Nikita and Da, Xingye and Ding, Runyu and Fan, Linxi and Fang, Yu and Fox, Dieter and Hu, Fengyuan and Huang, Spencer and others},
  journal={arXiv preprint arXiv:2503.14734},
  year={2025}
}

@article{jang2025dreamgen,
  title={DreamGen: Unlocking Generalization in Robot Learning through Video World Models},
  author={Jang, Joel and Ye, Seonghyeon and Lin, Zongyu and Xiang, Jiannan and Bjorck, Johan and Fang, Yu and Hu, Fengyuan and Huang, Spencer and Kundalia, Kaushil and Lin, Yen-Chen and others},
  journal={arXiv preprint arXiv:2505.12705},
  year={2025}
}

@article{zhu2025versatile,
  title={Versatile Loco-Manipulation through Flexible Interlimb Coordination},
  author={Zhu, Xinghao and Chen, Yuxin and Sun, Lingfeng and Niroui, Farzad and Cleac'h, Simon Le and Wang, Jiuguang and Fang, Kuan},
  journal={arXiv preprint arXiv:2506.07876},
  year={2025}
}

@article{mon2025embodied,
  title={Embodied large language models enable robots to complete complex tasks in unpredictable environments},
  author={Mon-Williams, Ruaridh and Li, Gen and Long, Ran and Du, Wenqian and Lucas, Christopher G},
  journal={Nature Machine Intelligence},
  pages={1--10},
  year={2025},
  publisher={Nature Publishing Group UK London}
}

@inproceedings{fu2023deep,
  title={Deep whole-body control: learning a unified policy for manipulation and locomotion},
  author={Fu, Zipeng and Cheng, Xuxin and Pathak, Deepak},
  booktitle={Conference on Robot Learning},
  pages={138--149},
  year={2023},
  organization={PMLR}
}

@inproceedings{zhi2025learning,
  title={Learning a Unified Policy for Position and Force Control in Legged Loco-Manipulation},
  author={Zhi, Peiyuan and Li, Peiyang and Yin, Jianqin and Jia, Baoxiong and Huang, Siyuan},
  booktitle={Conference on Robot Learning},
  pages={652--669},
  year={2025},
  organization={PMLR}
}

@inproceedings{chiu2022collision,
  title={A collision-free mpc for whole-body dynamic locomotion and manipulation},
  author={Chiu, Jia-Ruei and Sleiman, Jean-Pierre and Mittal, Mayank and Farshidian, Farbod and Hutter, Marco},
  booktitle={2022 international conference on robotics and automation (ICRA)},
  pages={4686--4693},
  year={2022},
  organization={IEEE}
}

@inproceedings{zimmermann2021go,
  title={Go fetch!-dynamic grasps using boston dynamics spot with external robotic arm},
  author={Zimmermann, Simon and Poranne, Roi and Coros, Stelian},
  booktitle={2021 IEEE International Conference on Robotics and Automation (ICRA)},
  pages={4488--4494},
  year={2021},
  organization={IEEE}
}

@article{ma2022combining,
  title={Combining learning-based locomotion policy with model-based manipulation for legged mobile manipulators},
  author={Ma, Yuntao and Farshidian, Farbod and Miki, Takahiro and Lee, Joonho and Hutter, Marco},
  journal={IEEE Robotics and Automation Letters},
  volume={7},
  number={2},
  pages={2377--2384},
  year={2022},
  publisher={IEEE}
}

@article{yang2025helom,
  title={HeLoM: Hierarchical Learning for Whole-Body Loco-Manipulation in Hexapod Robot},
  author={Yang, Xinrong and Li, Peizhuo and Li, Hongyi and Lu, Junkai and Chang, Linnan and Cao, Yuhong and Zhang, Yifeng and Sun, Ge and Sartoretti, Guillaume},
  journal={arXiv preprint arXiv:2509.23651},
  year={2025}
}

@inproceedings{portela2024learning,
  title={Learning force control for legged manipulation},
  author={Portela, Tifanny and Margolis, Gabriel B and Ji, Yandong and Agrawal, Pulkit},
  booktitle={2024 IEEE International Conference on Robotics and Automation (ICRA)},
  pages={15366--15372},
  year={2024},
  organization={IEEE}
}

@article{billard2019trends,
  title={Trends and challenges in robot manipulation},
  author={Billard, Aude and Kragic, Danica},
  journal={Science},
  volume={364},
  number={6446},
  pages={eaat8414},
  year={2019},
  publisher={American Association for the Advancement of Science}
}

@article{bai2025towards,
  title={Towards a unified understanding of robot manipulation: A comprehensive survey},
  author={Bai, Shuanghao and Song, Wenxuan and Chen, Jiayi and Ji, Yuheng and Zhong, Zhide and Yang, Jin and Zhao, Han and Zhou, Wanqi and Zhao, Wei and Li, Zhe and others},
  journal={arXiv preprint arXiv:2510.10903},
  year={2025}
}

@article{wang2025robot,
  title={Robot Manipulation Based on Embodied Visual Perception: A Survey},
  author={Wang, Sicheng and Nikoli{\'c}, Milutin N and Lam, Tin Lun and Gao, Qing and Ding, Runwei and Zhang, Tianwei},
  journal={CAAI Transactions on Intelligence Technology},
  year={2025},
  publisher={Wiley Online Library}
}

@article{gu2025humanoid,
  title={Humanoid locomotion and manipulation: Current progress and challenges in control, planning, and learning},
  author={Gu, Zhaoyuan and Li, Junheng and Shen, Wenlan and Yu, Wenhao and Xie, Zhaoming and McCrory, Stephen and Cheng, Xianyi and Shamsah, Abdulaziz and Griffin, Robert and Liu, C Karen and others},
  journal={arXiv preprint arXiv:2501.02116},
  year={2025}
}

@article{sleiman2023versatile,
  title={Versatile multicontact planning and control for legged loco-manipulation},
  author={Sleiman, Jean-Pierre and Farshidian, Farbod and Hutter, Marco},
  journal={Science Robotics},
  volume={8},
  number={81},
  pages={eadg5014},
  year={2023},
  publisher={American Association for the Advancement of Science}
}

@article{chitta2012mobile,
  title={Mobile manipulation in unstructured environments: Perception, planning, and execution},
  author={Chitta, Sachin and Jones, E Gil and Ciocarlie, Matei and Hsiao, Kaijen},
  journal={IEEE Robotics \& Automation Magazine},
  volume={19},
  number={2},
  pages={58--71},
  year={2012},
  publisher={IEEE}
}

@article{sandakalum2022motion,
  title={Motion planning for mobile manipulators—a systematic review},
  author={Sandakalum, Thushara and Ang Jr, Marcelo H},
  journal={Machines},
  volume={10},
  number={2},
  pages={97},
  year={2022},
  publisher={MDPI}
}

@inproceedings{sun2022fully,
  title={Fully autonomous real-world reinforcement learning with applications to mobile manipulation},
  author={Sun, Charles and Orbik, Jȩdrzej and Devin, Coline Manon and Yang, Brian H and Gupta, Abhishek and Berseth, Glen and Levine, Sergey},
  booktitle={Conference on Robot Learning},
  pages={308--319},
  year={2022},
  organization={PMLR}
}

@article{wu2023tidybot,
  title={Tidybot: Personalized robot assistance with large language models},
  author={Wu, Jimmy and Antonova, Rika and Kan, Adam and Lepert, Marion and Zeng, Andy and Song, Shuran and Bohg, Jeannette and Rusinkiewicz, Szymon and Funkhouser, Thomas},
  journal={Autonomous Robots},
  volume={47},
  number={8},
  pages={1087--1102},
  year={2023},
  publisher={Springer}
}

@inproceedings{jang2022bc,
  title={Bc-z: Zero-shot task generalization with robotic imitation learning},
  author={Jang, Eric and Irpan, Alex and Khansari, Mohi and Kappler, Daniel and Ebert, Frederik and Lynch, Corey and Levine, Sergey and Finn, Chelsea},
  booktitle={Conference on Robot Learning},
  pages={991--1002},
  year={2022},
  organization={PMLR}
}

@article{dmitry2018qt,
  title={Qt-opt. Scalable deep reinforcement learning for vision-based robotic manipulation},
  author={Dmitry, Kalashnikov and Alex, Irpan and Peter, Pastor and Julian, Ibarz and Alexander, Herzog and Eric, Jang and Deirdre, Quillen and Ethan, Holly and Mrinal, Kalakrishnan and Vincent, Vanhoucke and others},
  journal={arXiv preprint},
  year={2018}
}

@article{shafiullah2022behavior,
  title={Behavior transformers: Cloning $ k $ modes with one stone},
  author={Shafiullah, Nur Muhammad and Cui, Zichen and Altanzaya, Ariuntuya Arty and Pinto, Lerrel},
  journal={Advances in neural information processing systems},
  volume={35},
  pages={22955--22968},
  year={2022}
}

@article{liu2025diffusion,
  title={Diffusion Models in Robotics: A Survey},
  author={Liu, Xiaokang and Ma, Kevin Yuchen and Gao, Chen and Shou, Mike Zheng},
  year={2025}
}

@article{wolf2025diffusion,
  title={Diffusion models for robotic manipulation: A survey},
  author={Wolf, Rosa Petra and Shi, Yitian and Liu, Sheng and Rayyes, Rania},
  journal={Frontiers in Robotics and AI},
  volume={12},
  pages={1606247},
  year={2025},
  publisher={Frontiers}
}

@inproceedings{ze2023gnfactor,
  title={Gnfactor: Multi-task real robot learning with generalizable neural feature fields},
  author={Ze, Yanjie and Yan, Ge and Wu, Yueh-Hua and Macaluso, Annabella and Ge, Yuying and Ye, Jianglong and Hansen, Nicklas and Li, Li Erran and Wang, Xiaolong},
  booktitle={Conference on robot learning},
  pages={284--301},
  year={2023},
  organization={PMLR}
}

@article{dong2025m4diffuser,
  title={M4Diffuser: Multi-View Diffusion Policy with Manipulability-Aware Control for Robust Mobile Manipulation},
  author={Dong, Ju and Zhang, Lei and Zhang, Liding and Ling, Yao and Fu, Yu and Bai, Kaixin and M{\'a}rton, Zolt{\'a}n-Csaba and Bing, Zhenshan and Chen, Zhaopeng and Knoll, Alois Christian and others},
  journal={arXiv preprint arXiv:2509.14980},
  year={2025}
}

@article{gong2023legged,
  title={Legged robots for object manipulation: A review},
  author={Gong, Yifeng and Sun, Ge and Nair, Aditya and Bidwai, Aditya and CS, Raghuram and Grezmak, John and Sartoretti, Guillaume and Daltorio, Kathryn A},
  journal={Frontiers in Mechanical Engineering},
  volume={9},
  pages={1142421},
  year={2023},
  publisher={Frontiers Media SA}
}

@article{honerkamp2023n,
  title={{N$^{2}$M$^{2}$}: Learning navigation for arbitrary mobile manipulation motions in unseen and dynamic environments},
  author={Honerkamp, Daniel and Welschehold, Tim and Valada, Abhinav},
  journal={IEEE Transactions on Robotics},
  volume={39},
  number={5},
  pages={3601--3619},
  year={2023},
  publisher={IEEE}
}

@article{ha2024umi,
  title={Umi on legs: Making manipulation policies mobile with manipulation-centric whole-body controllers},
  author={Ha, Huy and Gao, Yihuai and Fu, Zipeng and Tan, Jie and Song, Shuran},
  journal={arXiv preprint arXiv:2407.10353},
  year={2024}
}

@article{barreiros2025careful,
  title={A careful examination of large behavior models for multitask dexterous manipulation},
  author={Barreiros, Jose and Beaulieu, Andrew and Bhat, Aditya and Cory, Rick and Cousineau, Eric and Dai, Hongkai and Fang, Ching-Hsin and Hashimoto, Kunimatsu and Irshad, Muhammad Zubair and Itkina, Masha and others},
  journal={arXiv preprint arXiv:2507.05331},
  year={2025}
}

@online{BostonDynamics_Atlas2025,
  author  = {Boston Dynamics},
  title   = {Large Behavior Models: Atlas Find New Footing},
  year    = {2025},
  month   = aug,
  url     = {https://bostondynamics.com/blog/large-behavior-models-atlas-find-new-footing/},
  urldate = {2025-11-29},
  date    = {2025-08-14}
}

@online{FigureAI_Helix2025,
  author  = {Figure AI},
  title   = {Helix: A Vision-Language-Action Model for Generalist Humanoid Control},
  date    = {2025-02-20},
  url     = {https://www.figure.ai/news/helix},
  urldate = {2025-11-29}
}

@article{intelligence2025pi05,
  title = {${\pi}_{0.5}$: A Vision--Language--Action Model with Open-World Generalization},
  author = {Intelligence, Physical and Black, Kevin and Brown, Noah and Darpinian, James and Dhabalia, Karan and Driess, Danny and Esmail, Adnan and Equi, Michael and Finn, Chelsea and Fusai, Niccolo and others},
  journal = {arXiv preprint arXiv:2504.16054},
  year = {2025}
}

@online{SundayRobotics_Technology2025,
  author  = {Sunday Robotics},
  title   = {Technology | Sunday Robotics},
  year    = {2025},
  url     = {https://www.sunday.ai/technology},
  urldate = {2025-11-29}
}

@online{Flexion_Reflect2025,
  author  = {Flexion Team},
  title   = {Flexion Reflect v0 - Towards Generalizable Robot Autonomy},
  date    = {2025-11-20},
  url     = {https://flexion.ai/news/flexion-reflect-v0-towards-generalizable-robot-autonomy},
  urldate = {2025-11-29}
}

@inproceedings{rehman2016towards,
  title={Towards a multi-legged mobile manipulator},
  author={Rehman, Bilal Ur and Focchi, Michele and Lee, Jinoh and Dallali, Houman and Caldwell, Darwin G and Semini, Claudio},
  booktitle={2016 IEEE international conference on robotics and automation (ICRA)},
  pages={3618--3624},
  year={2016},
  organization={IEEE}
}

@article{zhao2023learning,
  title={Learning fine-grained bimanual manipulation with low-cost hardware},
  author={Zhao, Tony Z and Kumar, Vikash and Levine, Sergey and Finn, Chelsea},
  journal={arXiv preprint arXiv:2304.13705},
  year={2023}
}

@inproceedings{kang2025incorporating,
  title={Incorporating task progress knowledge for subgoal generation in robotic manipulation through image edits},
  author={Kang, Xuhui and Kuo, Yen-Ling},
  booktitle={2025 IEEE/CVF Winter Conference on Applications of Computer Vision (WACV)},
  pages={7490--7499},
  year={2025},
  organization={IEEE}
}

@article{huang2024rekep,
  title={Rekep: Spatio-temporal reasoning of relational keypoint constraints for robotic manipulation},
  author={Huang, Wenlong and Wang, Chen and Li, Yunzhu and Zhang, Ruohan and Fei-Fei, Li},
  journal={arXiv preprint arXiv:2409.01652},
  year={2024}
}

@online{Krishnan2017LfDSurgical,
  author  = {Krishnan, Sanjay and Fox, Roy and Goldberg, Ken},
  title   = {Learning Long Duration Sequential Task Structure From Demonstrations with Application in Surgical Robotics},
  date    = {2017-10-17},
  url     = {https://bair.berkeley.edu/blog/2017/10/17/lfd-surgical-robots/},
  urldate = {2025-11-29}
}

@article{he2025demystifying,
  title={Demystifying Diffusion Policies: Action Memorization and Simple Lookup Table Alternatives},
  author={He, Chengyang and Liu, Xu and Camps, Gadiel Sznaier and Sartoretti, Guillaume and Schwager, Mac},
  journal={arXiv preprint arXiv:2505.05787},
  year={2025}
}

@article{oord2018representation,
  title={Representation learning with contrastive predictive coding},
  author={Oord, Aaron van den and Li, Yazhe and Vinyals, Oriol},
  journal={arXiv preprint arXiv:1807.03748},
  year={2018}
}

\end{document}